%% file: ICPR_2026_self_supervised/main.tex
\documentclass{llncs}
\usepackage[T1]{fontenc}
\usepackage{lmodern}     \usepackage{xcolor}
\usepackage[utf8]{inputenc}  

\usepackage{amsmath,amsfonts,amssymb}
\usepackage{booktabs}
\usepackage{subcaption}
\usepackage{adjustbox}

\usepackage{graphicx}
\usepackage{booktabs,tabularx,float}
\usepackage{caption}        
\usepackage{subcaption}

\usepackage{hyperref}
\hypersetup{
    colorlinks   = true,
    linkcolor    = blue,
    citecolor    = blue,
    urlcolor     = blue
}

\usepackage{tikz}
\usetikzlibrary{positioning,calc}

\usepackage{xcolor}

\input{ICPR_2026_self_supervised/preamble}

\begin{document}

\title{The Detector Teaches Itself: \\Lightweight Self-Supervised Adaptation for Open-Vocabulary Object Detection}

\author{Yazhe Wan \and
        Changjae Oh}
% \inst{1} \orcidID{0009-0005-1579-1049}
% \inst{1}\orcidID{1111-2222-3333-4444}

\authorrunning{Y. Wan and C. Oh}
\institute{Queen Mary University of London, London, UK\\
\email{\{yazhe.wan,c.oh\}@qmul.ac.uk}}

\maketitle

\input{ICPR_2026_self_supervised/content/abstract}
\input{ICPR_2026_self_supervised/content/introduction}

\input{ICPR_2026_self_supervised/content/related_work}
\input{ICPR_2026_self_supervised/content/method}

\input{ICPR_2026_self_supervised/content/experiment}
\input{ICPR_2026_self_supervised/content/conclusion}

\bibliographystyle{unsrt}
\bibliography{ICPR_2026_self_supervised/mybibfile} 
\end{document}

%% file: ICPR_2026_self_supervised/preamble.tex
\usepackage{array}        % 扩展列格式
\usepackage{tabularx}     % 自动伸缩列
\usepackage{booktabs}     % 专业横线
\usepackage{multirow}     % 跨行单元格
\usepackage{makecell}     
\usepackage{caption}      % 控制表题间距
\usepackage{float}        % [H] 固定位置
\usepackage{xcolor}        % 彩色支持
\usepackage{colortbl}      % 表格背景色
\usepackage{booktabs}      % 专业表格线条
\usepackage{array}         % 列格式控制
\usepackage{multirow}      % 多行表格（如需）

\renewcommand{\arraystretch}{1.25}     % 行高 1.25

\newcolumntype{L}[1]{>{\raggedright\arraybackslash}p{#1}}
\newcolumntype{C}[1]{>{\centering\arraybackslash}p{#1}}
\newcolumntype{R}[1]{>{\raggedleft\arraybackslash}p{#1}}
\newcolumntype{Y}{>{\centering\arraybackslash}X}   % tabularx 居中伸缩列

%% file: ICPR_2026_self_supervised/content/abstract.tex
\begin{abstract}

Open-vocabulary object detection aims to recognize objects from an open set of categories, which leverages vision-language models (VLMs) pre-trained on large-scale image-text data. The \emph{cooperative} paradigm combines an object detector with a VLM to achieve zero-shot recognition of novel objects. However, VLMs pre-trained on full images often struggle to capture local object details, limiting their effectiveness when applied to region-level detection. We present \emph{Decoupled Adaptivity Training (DAT)}, a self-supervised fine-tuning approach to improve VLMs for cooperative model-based object detection. Given a cooperative model consists of a closed-set detector and a VLM, we first construct a region-aware pseudo-labeled dataset using a pre-trained closed-set object detector, in which regions corresponding to novel objects may be present but remain unlabeled or mislabeled. We then fine-tune the visual backbone of the VLM in a decoupled manner, which enhances local feature alignment while preserving global semantic knowledge via weight interpolation. DAT is a plug-and-play module that requires no inference overhead and fine-tunes less than 0.8M parameters. Experiments on the COCO and LVIS datasets show that DAT consistently improves detection performance on both novel and known categories, establishing a new state of the art in cooperative open-vocabulary detection.

% Open-vocabulary object detection (OVD) aims to detect objects from an unbounded set of categories by leveraging pre-trained vision-language models (VLMs). A critical yet under-explored challenge is the \emph{global-local feature imbalance}: VLMs like SigLIP, pre-trained on holistic images, suffer from a significant performance drop when applied to cropped region proposals due to a domain shift, which severely compromises their local adaptivity. 

% To bridge this gap, we propose \emph{Decoupled Adaptivity Training (DAT)}, a novel and self-supervised fine-tuning paradigm. DAT introduces a self-contained adaptation loop where the detector teaches itself: it first constructs a region-aware dataset by harvesting object proposals and their pseudo-labels directly from a Mask R-CNN within a cooperative pipeline, eliminating the need for any external data. It then fine-tunes the visual backbone in a decoupled manner using a sigmoid-based contrastive loss to enhance local feature alignment, while strategically preserving global semantic knowledge via weight interpolation. This design makes DAT a plug-and-play module that introduces zero inference overhead. Extensive experiments on COCO and LVIS benchmarks demonstrate that DAT significantly boosts novel class AP (e.g., achieving 70.1 AP on COCO novel classes) without compromising base class performance. Remarkably, this gain is achieved by fine-tuning less than 0.8M parameters in under two hours on a single GPU which establishes a new state-of-the-art in efficiency and effectiveness for OVD adaptation.

\vspace{\baselineskip}
\textbf{Keywords:} Open-vocabulary object detection, parameter-efficient domain adaptation, self-supervised learning, vision-language models
\end{abstract}

%% file: ICPR_2026_self_supervised/content/introduction.tex
\section{Introduction}

\begingroup
\renewcommand{\thefootnote}{}
\footnotetext{The code and more results are available at: \url{https://qm-ipalab.github.io/DAT/}}
\endgroup
Detecting and recognizing objects beyond a fixed set of categories is a central challenge in computer vision, with applications in autonomous navigation, robotic perception, and surveillance systems. Traditional closed-set object detectors are limited to detecting objects within a predefined set of labels, which is unsuitable for real-world scenarios where new objects constantly appear~\cite{zheng2022towards}. Open-Vocabulary Object Detection (OVD) bridges this gap by leveraging language supervision to localize and recognize arbitrary objects~\cite{wang2023open}.

\begin{figure*}[t]
    \centering
    \includegraphics[width=\linewidth]{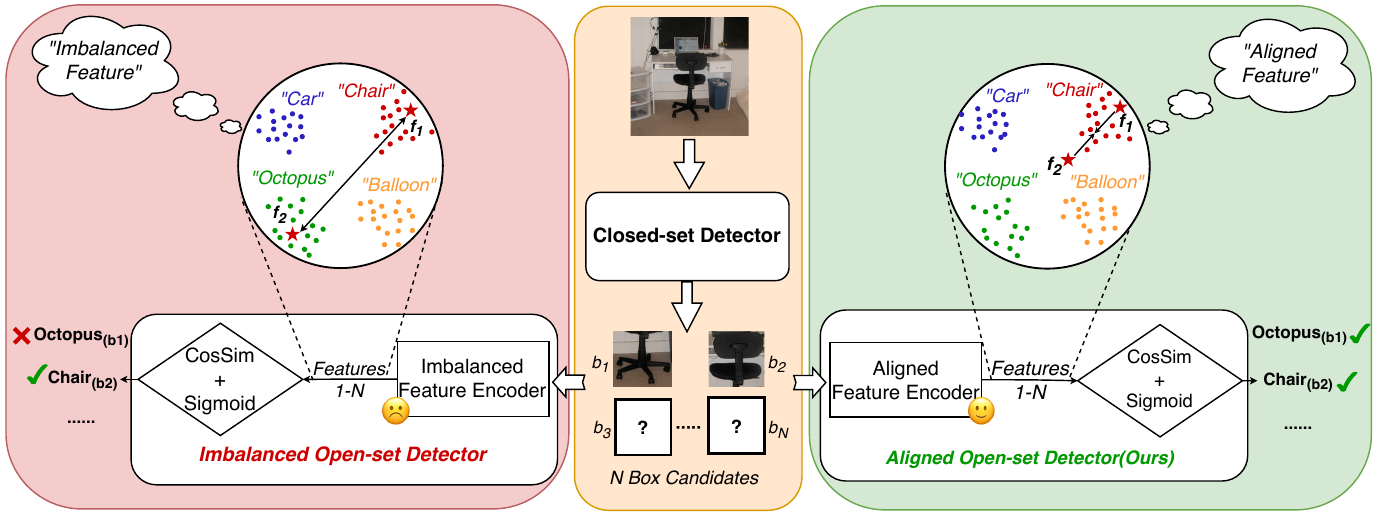}
    \caption{Global-local feature imbalance in VLMs under cooperative closed-set and open-set detection. Existing methods (left) leverage VLMs for open-vocabulary detection while overlooking the VLMs' limited capability in encoding region-level representations. Our approach (right) improves local adaptivity of VLMs without sacrificing the generalization capability.}
    \label{fig:pipeline}
\end{figure*}

Vision-Language Models (VLMs), such as CLIP~\cite{radford2021learning} and SigLIP~\cite{zhai2023sigmoid}, have enabled significant advancement in OVD. These foundation models are trained on vast amounts of image-text pairs, learning a shared space where visual and textual concepts can be directly compared. This zero-shot capability is ideal for OVD in principle, but integrating it into detection systems remains challenging. Existing approaches can be grouped into three categories~\cite{zareian2021open,gu2021open}: 
\textit{1) knowledge distillation}, where a detector learns from soft labels generated by a frozen VLM~\cite{kuoopen}; 
\textit{2) region-text alignment}, which explicitly aligns visual region features with text embeddings using region-caption data~\cite{li2025unbiased}; and 
\textit{3) region-text pre-training}, where models are trained end-to-end on region-text pairs~\cite{zhong2022regionclip}. 
However, these methods suffer from a domain gap between holistic images used for VLM pre-training and cropped object regions used during detection.

This domain gap is caused by an imbalance between global and local features. That is, VLMs excel at global image-level understanding, but struggle with cropped (local) regions, which may include partially visible or less attentive objects. VLMs show a noticeable drop in performance on novel categories when directly applied to regions~\cite{minderer2023scaling,bangalath2022bridging}. This further emphasizes the need for efficient adaptation methods that enhance regional understanding without losing global knowledge or requiring excessive computation.

Existing adaptation methods rely on external resources such as manual captions~\cite{yaofilip} or complex multi-model setups~\cite{gu2021zero}, which can introduce noise and increase complexity. While a self-supervised method does not require extra resources but efficiently allows the model to teach itself with own outputs as training signals. In particular, using region proposals from a detector like Mask R-CNN~\cite{he2017mask} to create region-text pairs is appealing, as it aligns the training data with the target task without need for extra annotations~\cite{du2022learning}. However, a key question remains: how can we enhance regional feature adaptivity of VLMs while preserving the model’s zero-shot ability in a simple and self-contained way.

We propose Decoupled Adaptivity Training (DAT) as illustrated in Fig.\ref{fig:pipeline}, a self-supervised fine-tuning method that resolves the imbalance of global-local features exists in VLMs for OVD by boosting the adpativity of VLM itself. Our approach is based on two ideas. First, we adapt only the visual encoder using a contrastive loss suited for multi-label regions, while regularizing the model to retain its original global knowledge. Second, we show that the detector itself can provide the necessary supervision. DAT uses cropped regions and pseudo-labels from a Mask R-CNN detector within a standard OVD pipeline, creating a self-training loop where the detector teaches the VLM. This eliminates the need for external data and ensures perfect alignment with the target domain. DAT modifies only the visual backbone, making it architecture-agnostic and free of inference overhead.

Our main contributions are summarized as follows:
\begin{itemize}
   
    \item We propose DAT, a self-supervised fine-tuning strategy that improves region-level alignment using a sigmoid-based contrastive loss and preserves global knowledge via weight interpolation.
    \item We design a self-contained adaptation loop that builds a region-aware dataset from the detector’s own pseudo-labels, avoiding external annotations and ensuring domain alignment.
    \item We validate DAT on COCO~\cite{lin2014microsoft} and LVIS~\cite{gupta2019lvis}, demonstrating that DAT improves novel class detection while maintaining base class performance.
\end{itemize}

%% file: ICPR_2026_self_supervised/content/related_work.tex
\section{Related Work}

\subsection{Open-vocabulary object detection}\label{subsec:OVD}
OVD aims to recognize objects beyond predefined categories, typically by leveraging VLMs pre-trained on large-scale image-text pairs. 
End-to-end methods train a unified model from region-text data. OV-DETR~\cite{zang2022open} adapts the DETR architecture~\cite{carion2020end} with text-conditioned matching, while Grounding DINO~\cite{liu2024grounding} combines a language-aware transformer with a denoising training strategy to improve open-world localization. DetCLIPv3~\cite{yao2024detclipv3} scales up this paradigm by jointly training on millions of region-text pairs.

Cooperative models, on the other hand, decouple the detector and the VLM, using the former to propose regions and the latter to classify them. ViLD~\cite{gu2021open} distills knowledge from CLIP into a detector, while Detic~\cite{zhou2022detecting} directly replaces classifier weights with CLIP embeddings. To improve region-VLM alignment,  BARON~\cite{wu2023aligning} introduces a region proposal network and elaborated feature fusion, while CORA~\cite{wu2023cora} uses cross-attention and further refines this with caption-aided training to associate regions with text concepts. Cooperative Foundational Models (CFM)~\cite{bharadwaj2025enhancing} combine Grounding DINO~\cite{liu2024grounding}, SAM~\cite{kirillov2023segment}, and CLIP to leverage their complementary strengths. Cooperative models generally improves the performance of a single OVD, but VLMs pre-trained on full images may have suboptimal performance in detecting objects within local image regions.

% These observations motivate our work. Rather than pursuing a unified region-text model or a detached distillation pipeline, we focus on enhancing the region-level recognition ability of the VLM itself within a cooperative detection framework. Our goal is to adapt the VLM to better understand localized visual content while preserving its inherent open-vocabulary strength—effectively bridging the gap between holistic pre-training and region-based detection.

\subsection{Parameter-efficient fine-tuning}
Parameter-Efficient Fine-Tuning (PEFT) is a crucial paradigm for adapting VLMs to specific domains while preserving their general capabilities~\cite{hetowards} to mitigate substantial computational requirements and the risk of catastrophic forgetting.
WiSE-FT~\cite{wortsman2022robust} employs weighted interpolation between pre-trained and fine-tuned weights to balance specialization and generalization. This approach demonstrates that carefully blending original and adapted parameters can maintain strong zero-shot performance while improving task-specific accuracy. LoRA~\cite{hu2022lora} introduces low-rank adapters that modify only a small subset of model parameters, achieving comparable performance to full fine-tuning with significantly reduced memory footprint.
% The field has also witnessed architectural innovations aimed at improving fusion efficiency in vision-language tasks. MambaOVD~\cite{liu2025mambaovd} replaces traditional transformer-based fusion modules with a Mamba architecture to reduce computational overhead, while OmDet-Turbo~\cite{zhao2024real} introduces an Efficient Fusion Head to alleviate computational bottlenecks during region-text alignment. In a different direction, SIA-OVD~\cite{wang2024sia} proposes a Shape-Invariant Adapter to address feature map deformation during region extraction without updating core CLIP parameters.
In VLMs, Visual Prompt Tuning~\cite{jia2022visual} learns task-specific prompts within the input embedding space, while Adapter-based approaches like CLIP-Adapter~\cite{gao2024clip} insert small learnable modules between transformer layers. Recent work on Diffusion-based adaptation~\cite{zhang2025diffclip} explores using diffusion models to generate synthetic training data for VLM fine-tuning.

These existing PEFT methods mainly focus on modifying model architecture or fusion mechanisms~\cite{chen2022adaptformer}. Moreover, many approaches still require external labeled data or complex training procedures. 
In our framework, we focus on adapting the visual encoder's feature representations through targeted fine-tuning with self-generated pseudo-labels, offering a solution that is architecturally simple, data-efficient, and highly effective at bridging the global-local feature gap.

% These diverse strategies collectively advance the efficiency of VLM adaptation. However, most existing PEFT methods focus on modifying model architecture or fusion mechanisms~\cite{chen2022adaptformer}, while few directly address the fundamental feature representation mismatch at the visual encoder level. Moreover, many approaches still require external labeled data or complex training procedures. Our DAT framework takes a complementary path by adapting the visual encoder's feature representations through targeted fine-tuning with self-generated pseudo-labels, offering a solution that is architecturally simple, data-efficient, and highly effective at bridging the global-local feature gap.

\subsection{Self-supervised learning and data efficiency}
The success of deep learning models often depends on large amounts of annotated data, which can be prohibitively expensive and time-consuming to acquire for specialized domains. Self-supervised learning offers a compelling alternative by generating supervisory signals from unlabeled data, making model adaptation more scalable and cost-effective~\cite{jing2020self}.

In the context of OVD, several strategies have been presented for self-supervised fine-tuning of VLMs. Caption-based supervision~\cite{desai2021virtex} leverages image captions where textual descriptions provide weak supervision for visual concepts, but these often lack precise alignment with specific object regions. More direct approaches utilize detector proposals~\cite{wang2020frustratingly}, which employ existing object detectors to generate region proposals and class predictions that serve as pseudo-labels for training. Recent advancements include techniques employing self-posed visual questions~\cite{sun2024sq}, where large language models automatically generate informative image-text pairs to create synthetic training data. In our work, we employ pseudo detection labels to fine-tune VLMs in a self-supervised manner.

%% file: ICPR_2026_self_supervised/content/method.tex
\section{Methodology}\label{sec:method}

\begin{figure*}[t]
    \centering
    \includegraphics[width=\linewidth]{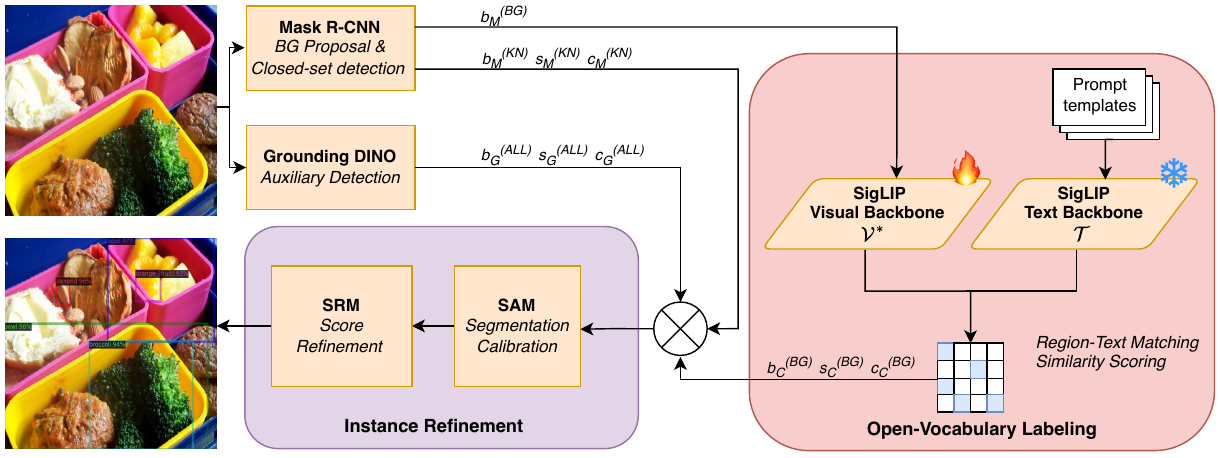}
    \caption{The Cooperative Foundational Models (CFM) pipeline with the VLM (CLIP) visual backbone fine-tuned with our decoupled adaptivity training (DAT). CFM synergizes an auxiliary detector (Grounding DINO) with a closed-set detector (Mask R-CNN), where its \textit{background} proposed bounding boxes ($b_M^{(BG)}$) may include objects with novel categories and hence can be used as input for open-vocabulary labeling by SigLIP. Segment Anything model (SAM) and the score refinement module (SRM) then refine the detection results based on the estimated boxes ($b$) and their associated scores ($s$) and classes($c$). DAT enhances this pipeline by adapting the VLM visual backbone to address the global-local feature imbalance.}
    \label{fig:coop}
\end{figure*}

We first outline the overall cooperative pipeline (Sec.~\ref{subsec:CFM}), and then present the proposed DAT, which refines the VLM's ability to process region proposals (Sec.~\ref{subsec:DAT}). Finally, we introduce the optimization and implementation details (Sec.~\ref{subsec:opt}).

\subsection{Preliminary: Cooperative Foundational Models}\label{subsec:CFM}
Fig.~\ref{fig:coop} shows the pipeline of CFM~\cite{bharadwaj2025enhancing} that combines the strengths of multiple vision models. CFM leverages the complementary abilities of different models: \textit{an open-set detector}~\cite{liu2024grounding}, which is capable of zero-shot detection, \textit{a closed-set detector}~\cite{he2017mask}, which precisely localizes seen categories, and \textit{a VLM}~\cite{radford2021learning}, which provides further open-vocabulary capability to recognize the unrecognized instances that are implicitly classified as background by the closed-set detector. 

Specifically, a set of learnable prompt templates (e.g., “a photo of a [CLASS]”) are applied and then encoded by SigLIP’s text encoder to generate class-specific embeddings. These embeddings are matched with visual features from cropped regions classified as \textit{background} by the closed-set detector, assuming that the closed-set detector has misclassified some regions as background. The final detections are obtained by aggregating and refining the outputs from these components through confidence-based fusion and non-maximum suppression.

While this cooperative design is effective in identifying objects in a complementary manner, we observe that the \textit{plug-and-play} use of SigLIP may not be suitable for object detection. The SigLIP model, pre-trained on holistic images with their text prompts, is directly applied to cropped regions without adaptation, and hence SigLIP may suffer from encoding features of local regions. With the CFM pipeline shown in Fig.~\ref{fig:coop}, the proposed DAT aims to tackle this limitation by carefully fine-tuning the SigLIP visual backbone to enhance its region-level understanding without altering the cooperative inference process.

\subsection{Decoupled Adaptivity Training (DAT)}\label{subsec:DAT}

\begin{figure*}[t]
    \centering
    \includegraphics[width=0.8\linewidth]{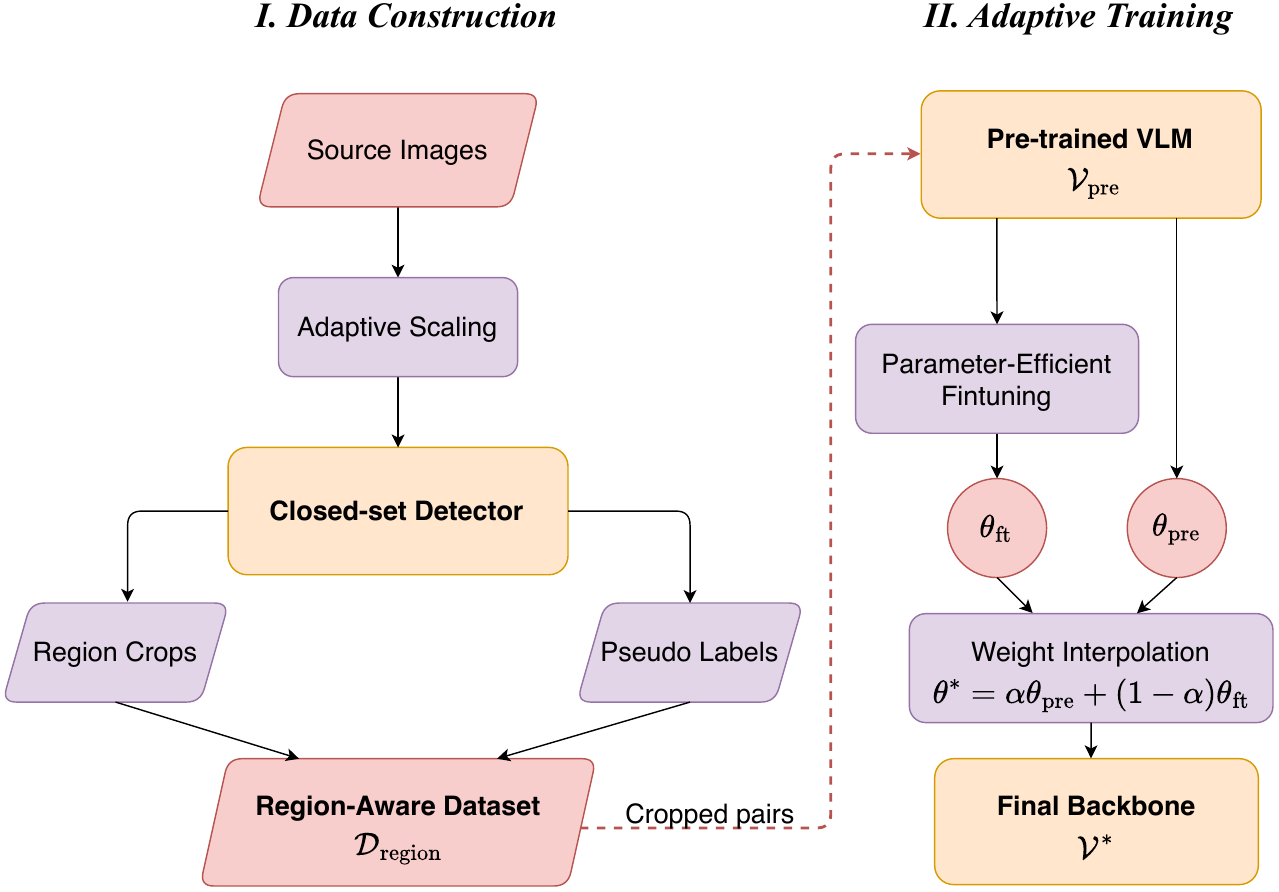}
    \caption{Overview of the proposed Decoupled Adaptivity Training (DAT). Using the region-award dataset, the pre-trained VLM is fine-tuned in a self-supervised and parameter-efficient manner, and the final backbone is integrated into the detection pipeline.}
    \label{fig:DAT}
\end{figure*}

Fig.~\ref{fig:DAT} shows the overall pipeline of DAT, which fine-tunes the SigLIP visual backbone in Fig.~\ref{fig:coop} in a decoupled manner.
Given the pre-trained visual backbone $\mathcal{V}_{\text{pre}}$, we transform $\mathcal{V}{\text{pre}}$ into a region-adapted model $\mathcal{V}^{*}$ through a streamlined adaptation pipeline. We first construct the region-aware dataset $\mathcal{D}_{\text{region}}$ by using a closed-set detector (Mask R-CNN) to use their outputs as the pseudo-labeled detection dataset.
To balance the scale and quality of the constructed dataset, we introduce two key hyperparameters: $N_{\text{max}}$, the maximum number of box candidates retained per image, and $\tau_{\text{conf}}$, the confidence score to keep high-quality proposals only. For instance, this configuration yields approximately 400 cropped regions (with predicted labels) from 5K source images. 
% Each region is resized to $224\times224$ pixels and paired with its corresponding class name—derived from either pseudo-labels or ground-truth annotations—to form region-level image–text pairs that closely match the target region-of-interest (RoI) distribution encountered during inference.

During the DAT stage, we fine-tune $\mathcal{V}_{\text{pre}}$ on $\mathcal{D}_{\text{region}}$ using a region-level contrastive objective designed to strengthen the alignment between visual regions and their textual descriptions. The \textit{decoupled} nature of our approach operates on two distinct levels. First, we decouple the VLM from the pipeline to avoid the need for costly end-to-end fine-tuning of the entire detection system. By freezing the detector and focusing on the adaptation of vision-language model (VLM), we maintain the quality of the proposals while enabling rapid and modular adaptation of the recognition component. Second, and more critically, we decouple the visual backbone from the full VLM during fine-tuning. The semantic alignment capabilities of VLMs are primarily embedded in their visual encoder, while the textual backbone is more responsible for generalizability across domains. By isolating and adapting only the visual encoder $\mathcal{V}_{\text{pre}}$, we can substantially reduce the number of trainable parameters, making fine-tuning efficient and scalable, and maintain the generalization capacity of textual embeddings. 

% In this case, we allow the visual encoder to specialize in region-level representation without disrupting the cross-modal alignment mechanisms learned during pre-training. 

Furthermore, to preserve the VLM's original zero-shot capability while enhancing its local discriminability, we adopt WiSE-FT~\cite{wortsman2022robust}, which interpolates between the pre-trained and fine-tuned model weights. This adaptive interpolation prevents catastrophic forgetting of global semantic knowledge while enabling effective region-level adaptation. As a result, $\mathcal{V}^{*}$ not only improves its sensitivity to local object features but also maintains strong generalization to unseen categories. Importantly, this adaptation not only enhances the VLM's ability to detect novel objects but also helps recover base-category instances that the Mask R-CNN detector may misclassify due to its architectural limitations.

Finally, the adapted backbone $\mathcal{V}^{*}$ directly replaces the original $\mathcal{V}_{\text{pre}}$ in the cooperative pipeline, requiring no structural changes or additional inference cost.

\subsection{Optimization and implementation details}\label{subsec:opt}

% DAT performs a decoupled fine-tuning process that only updates the vision encoder
% This design allows the model to learn localized semantic adaptivity independently from detection heads or language components, making the adapted backbone $\mathcal{V}^{*}$ directly interchangeable within the cooperative framework.

Given a cropped image and its corresponding text label, we measure the similarity between the visual and textual embeddings, defined as $z$. 
We employ the Sigmoid Contrastive loss:
\begin{equation}
\mathcal{L}_{\text{adapt}}=-\frac{1}{B}\sum_{i=1}^{B}\Bigl[y_{i}\log\sigma(z)+(1-y_{i})\log(1-\sigma(z))\Bigr],
\label{eq:adapt_loss}
\end{equation}
where $y_{i}\in\{0,1\}$ indicates whether the image–text pair is positive or negative.  
Compared with the softmax-based CLIP loss, this sigmoid formulation accommodates multi-label and partially overlapping RoIs, improving robustness in dense visual regions.

To retain the strong global semantics of the pre-trained SigLIP model, we adopt WiSE-FT~\cite{wortsman2022robust} as a global stability constraint. After fine-tuning, the final adapted weights $\theta^{*}$ are computed via linear interpolation:
\begin{equation}
\theta^{*}=\alpha\theta_{\text{pre}}+(1-\alpha)\theta_{\text{ft}},\quad\alpha\in[0,1].
\label{eq:wise_ft}
\end{equation}
This interpolation blends the generalization capacity of $\theta_{\text{pre}}$ with the local specialization of $\theta_{\text{ft}}$, balancing generalizability and adaptivity. We empirically set $\alpha=0.5$ and the interpolation is applied once post-training.
% and requires less than 10 seconds on CPU, introducing no runtime cost.

% \paragraph{Adaptive Property.}
% The “adaptive” nature of DAT emerges from two complementary mechanisms:
% (1) selective parameter optimization—only the final LayerNorm and projection weights are updated, enabling targeted adaptation within a compact parameter subspace; and
% (2) adaptive subspace blending via Wise-FT, which dynamically interpolates global and local representations. 
% Together, these mechanisms enable efficient, self-regularizing adaptation that scales gracefully to diverse region-level distributions.

\textbf{Implementation details.} 
% \paragraph{Backbone and Trainable Parameters.}
We employ ViT-SO400M-14-SigLIP~\cite{kaaser2024dynamic} as the visual backbone, freezing all layers except the final LayerNorm and projection weights (approximately 0.8M trainable parameters). This selective fine-tuning scheme minimizes overfitting risk and ensures compatibility with the cooperative pipeline.
% \paragraph{Single-GPU Training on RTX 5090.}
Cropped images are generated by Mask R-CNN with a maximum of 80 regions per image, with a confidence threshold of 0.7.
We experiment with four dataset scales, 100, 500, 1000, and 5000 source images, yielding about 8K, 40K, 80K, and 400K crops, respectively. 
All crops are resized to $224\times224$ and trained with a batch size of 64, an initial learning rate of $1\times10^{-5}$, cosine learning-rate decay, and 5 epochs. The resulting backbone $\mathcal{V}^{*}$ can be seamlessly integrated into the cooperative pipeline, enabling improved region-level performance without any inference-time overhead.  
All code and trained weights will be released.

%% file: ICPR_2026_self_supervised/content/experiment.tex
\section{Experiments}\label{sec:exp}

We conducted comprehensive experiments to validate the effectiveness and efficiency of our proposed lightweight region-level adaptation strategy based on the SigLIP backbone~\cite{zhai2023sigmoid}. 
All training experiments were performed using a single RTX~5090 GPU without multi-GPU acceleration or distributed training. 
After fine-tuning, the adapted visual backbone is directly deployed into the CFM pipeline by replacing the original vision encoder, without any structural or inference-time modification.

\input{ICPR_2026_self_supervised/tables/tab_coco}
\subsection{Setup}\label{subsec:data}
In the training phase, we fine-tuned SigLIP using the COCO~2017 Train split~\cite{lin2014microsoft} through our proposed DAT framework. Following the standard 48/17 open-vocabulary split established in~\cite{zareian2021open} and adopted in subsequent works~\cite{gu2021open,zhou2022detecting,wu2023aligning,wu2023cora}, we use the 48 base categories for training and reserve the remaining 17 novel categories for evaluation.

To enable self-supervised region adaptation, we integrate a Mask R-CNN detector that has been pre-trained only on the base categories. As Mask R-CNN is a closed-set detector that cannot recognize or propose regions of unseen classes, the fine-tuning process naturally ensures that proposed regions corresponding to novel objects may be present but remain unlabeled or mislabeled to SigLIP. During this stage, we set the data-generation hyperparameters $N_{\text{max}}$ (maximum proposals per image) and $\tau_{\text{conf}}$ (confidence threshold) to 80 and 0.7 respectively, balancing data quantity with annotation quality. This configuration yields approximately 400K high-quality image--annotation pairs. Such a strict setup guarantees that the evaluation faithfully reflects the model's ability to generalize to novel object categories without any contamination from training-time exposure.

We conduct a comprehensive evaluation of our method on two standard open-vocabulary object detection benchmarks: COCO 2017 OVD~\cite{zareian2021open}, which contains both seen and unseen categories in a 48/17 split, and LVIS v1.0~\cite{gupta2019lvis} for large-vocabulary generalization, where we follow common practice~\cite{zhou2022detecting,bharadwaj2025enhancing} and treat its 80 seen and 1123 unseen categories as the open-vocabulary evaluation set. Both evaluations adhere to established protocols to ensure fair and reproducible comparison with prior work.

\input{ICPR_2026_self_supervised/tables/tab_lvis}

We report COCO-style Average Precision (AP) across multiple IoU thresholds: {AP} (averaged over IoU $\in$ [0.50:0.95]), {AP50} (IoU=0.50), and {AP75} (IoU=0.75). Following the convention in open-vocabulary detection literature, we separately report performance on base and novel categories to clearly distinguish between in-distribution and out-of-distribution generalization. All metrics are computed using the official COCO and LVIS evaluation toolkits.

Our evaluation includes a representative set of state-of-the-art methods spanning different OVD paradigms: (1) End-to-end detection models trained with region-text alignment: OV-DETR~\cite{zang2022open} and DetCLIPv3~\cite{yao2024detclipv3} (2)Cooperative detection frameworks that decouple region proposal and classification: ViLD~\cite{gu2021open}, Detic~\cite{zhou2022detecting}, BARON~\cite{wu2023aligning}, CORA/CORA+~\cite{wu2023cora} and the CFM~\cite{bharadwaj2025enhancing}.
We compare using their officially reported numbers or re-implemented results under the same evaluation protocol. All compared methods are evaluated on the same dataset splits and metrics.

In our pipeline, we use Mask R-CNN (ResNet-101-FPN)~\cite{he2017mask} as the closed-set detector, Grounding DINO~\cite{liu2024grounding} as the open-set detector, ViT-H SAM~\cite{kirillov2023segment} as segmentations calibrator and ViT-SO400M-14-SigLIP as the base VLM. For all main results reported in Tables~\ref{tab:main_results} and~\ref{tab:lvis_results}, the SigLIP backbone is fine-tuned on 1,000 source images ($\approx$80k cropped regions), which is the optimal configuration identified in Table~\ref{tab:abl_dataset_scale}.

\subsection{Results}\label{subsec:main}
Table~\ref{tab:main_results} summarizes the results on the COCO~2017 OVD split, and Table~\ref{tab:lvis_results} presents the corresponding results on LVIS~v1.  
Our fine-tuned SigLIP backbone achieves the best performance within the cooperative detection paradigm in novel categories, while maintaining competitive performance in base categories.  
These results indicate that targeted adaptation of high-level visual embeddings—without modifying the full backbone—is sufficient to enhance region-text alignment.

\begin{figure}[!t]
    \centering
    \scalebox{0.93}{
    \begin{tabular}{c c c c c}
        \includegraphics[width=0.2\textwidth]{ICPR_2026_self_supervised/figure/Raw/r1.jpeg} &
        \includegraphics[width=0.2\textwidth]{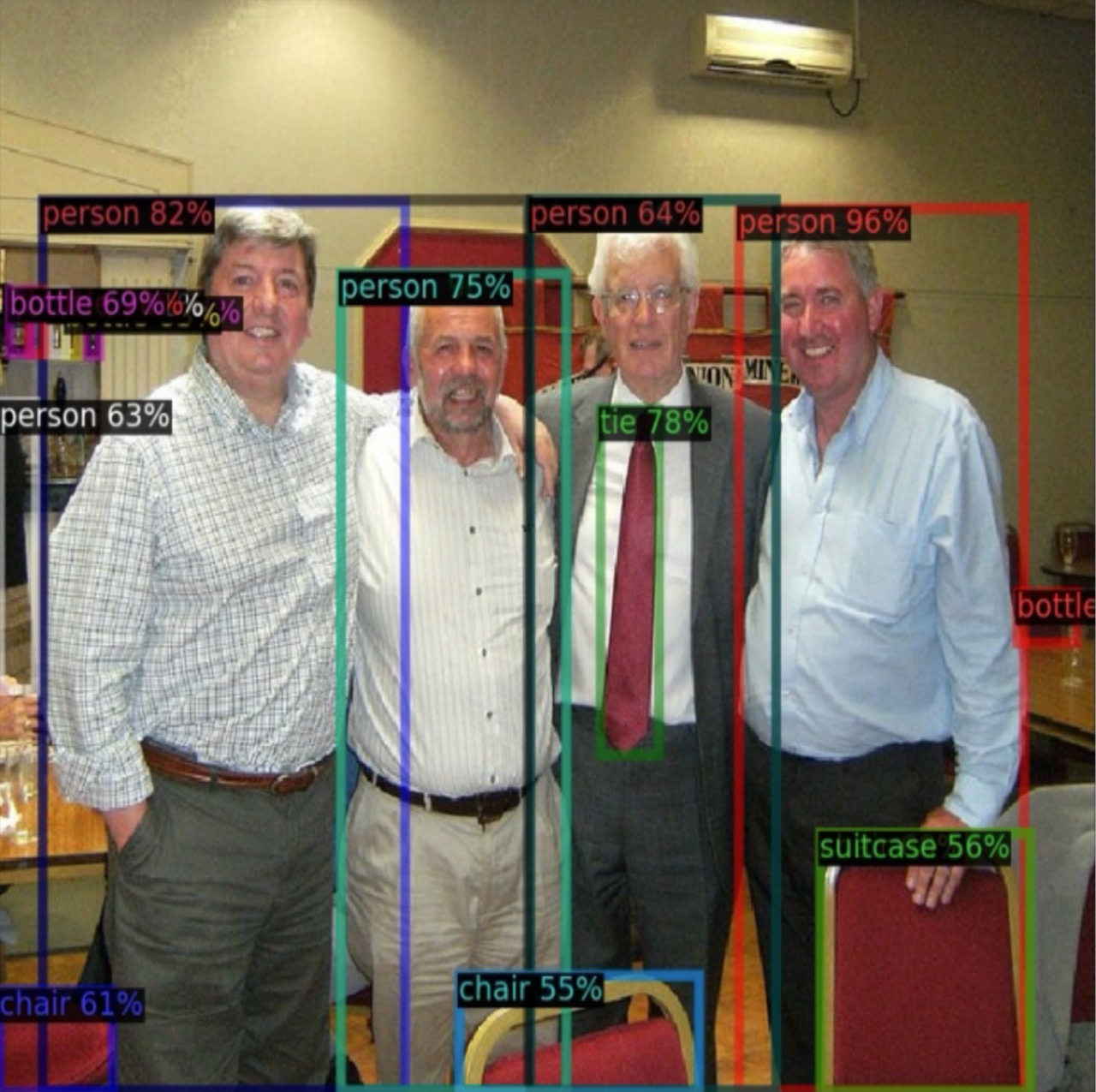} &
        \includegraphics[width=0.2\textwidth]{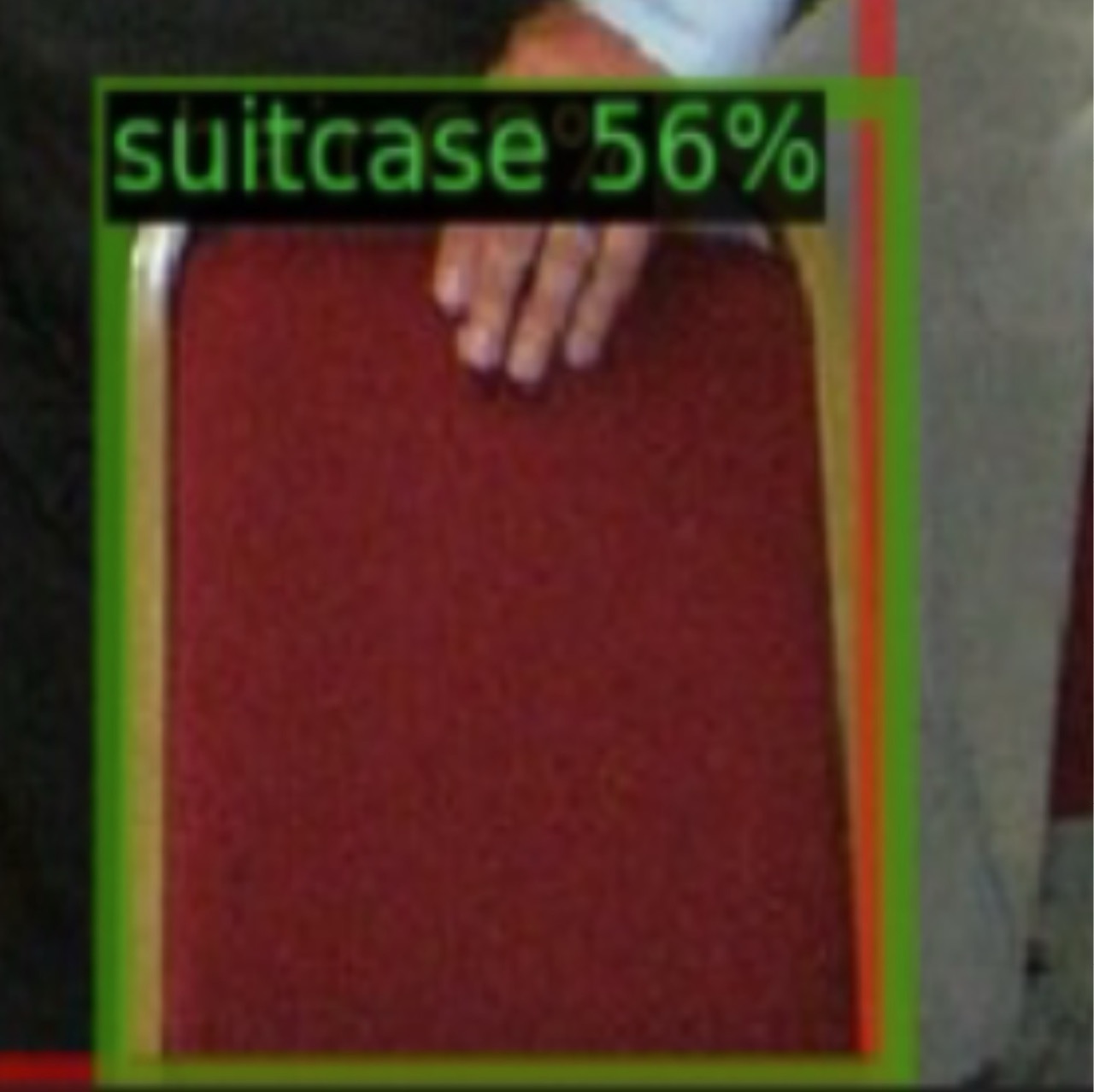} &
        \includegraphics[width=0.2\textwidth]{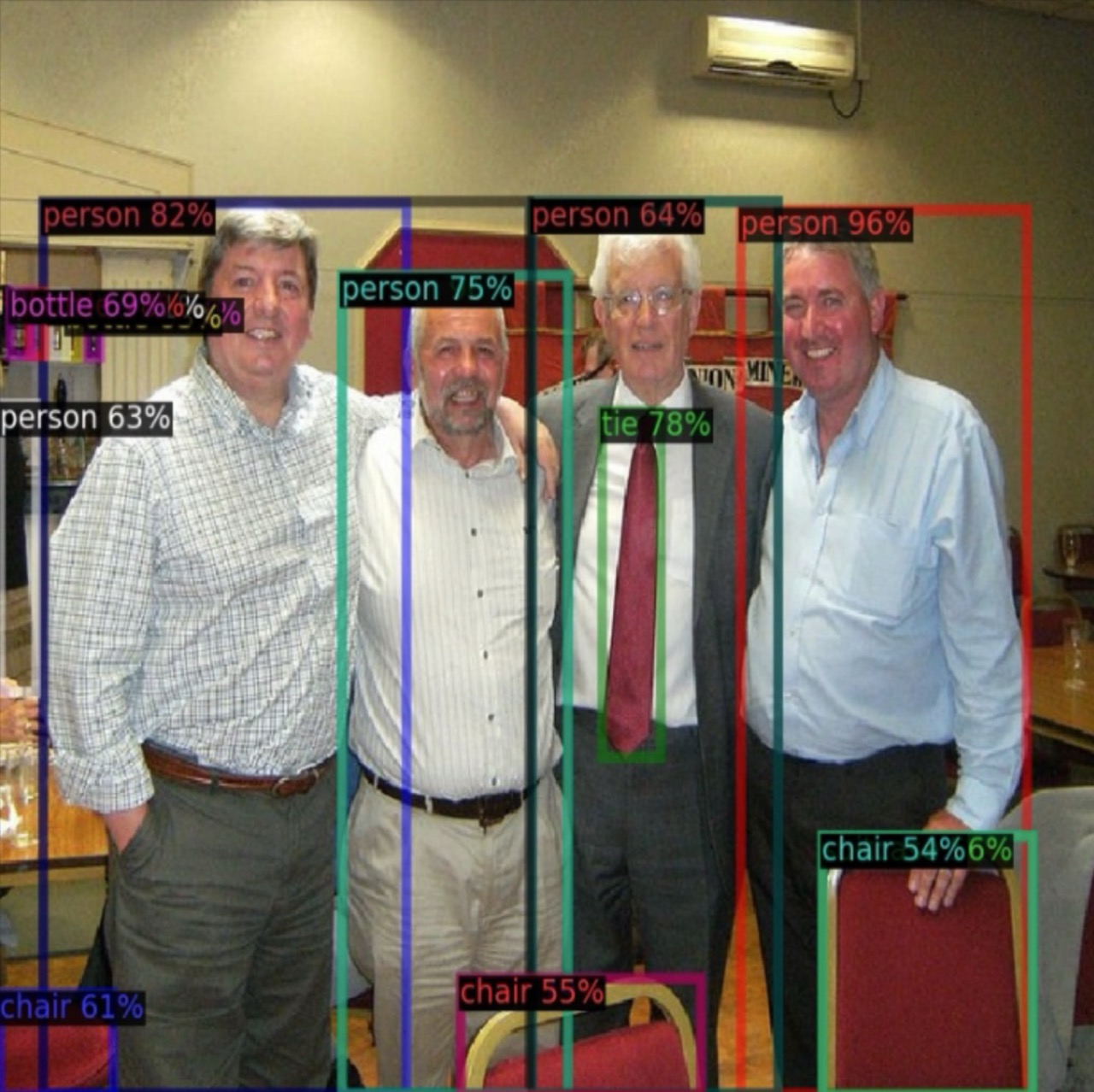} &
        \includegraphics[width=0.2\textwidth]{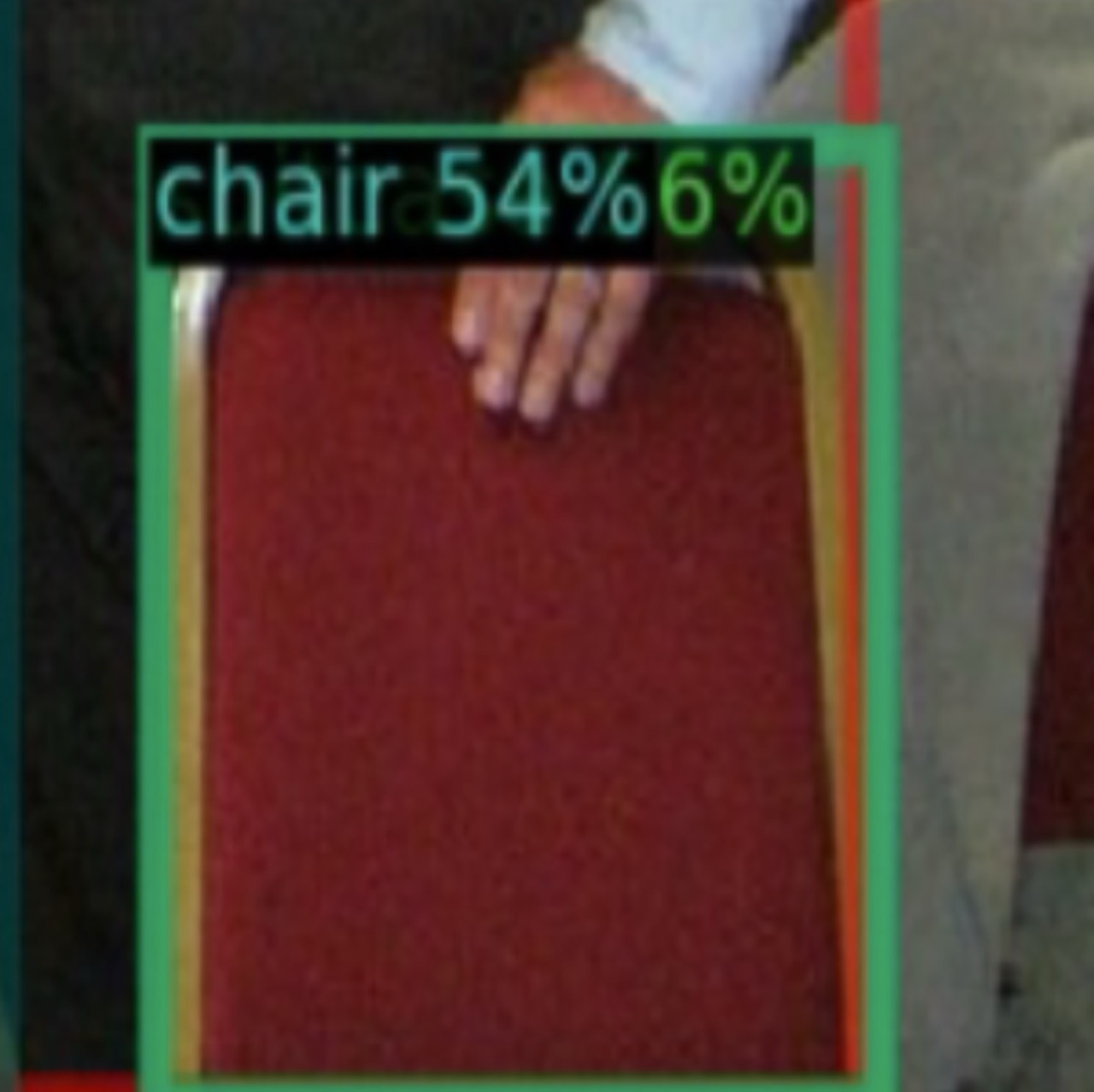} \\
        \includegraphics[width=0.2\textwidth]{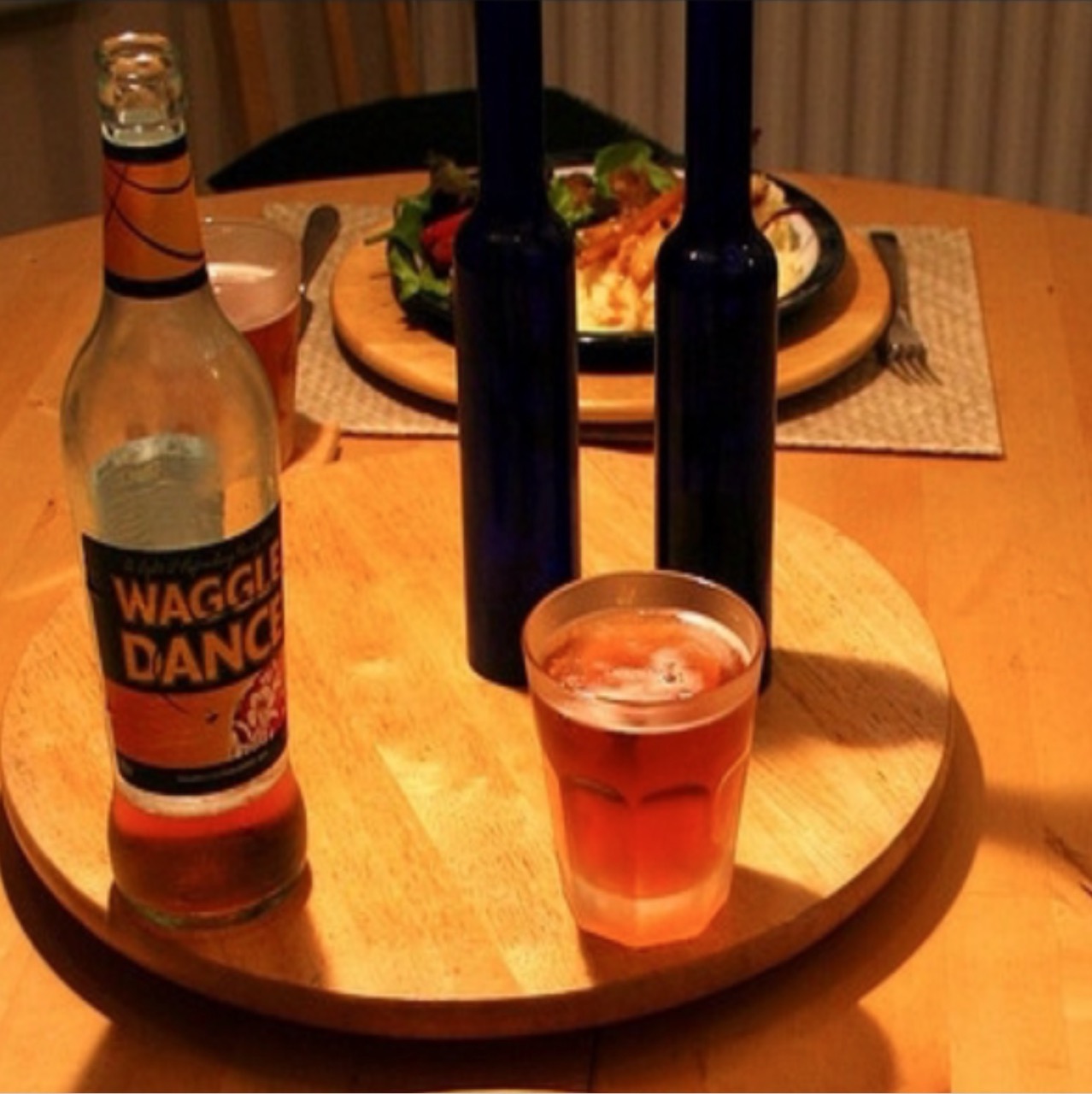} &
        \includegraphics[width=0.2\textwidth]{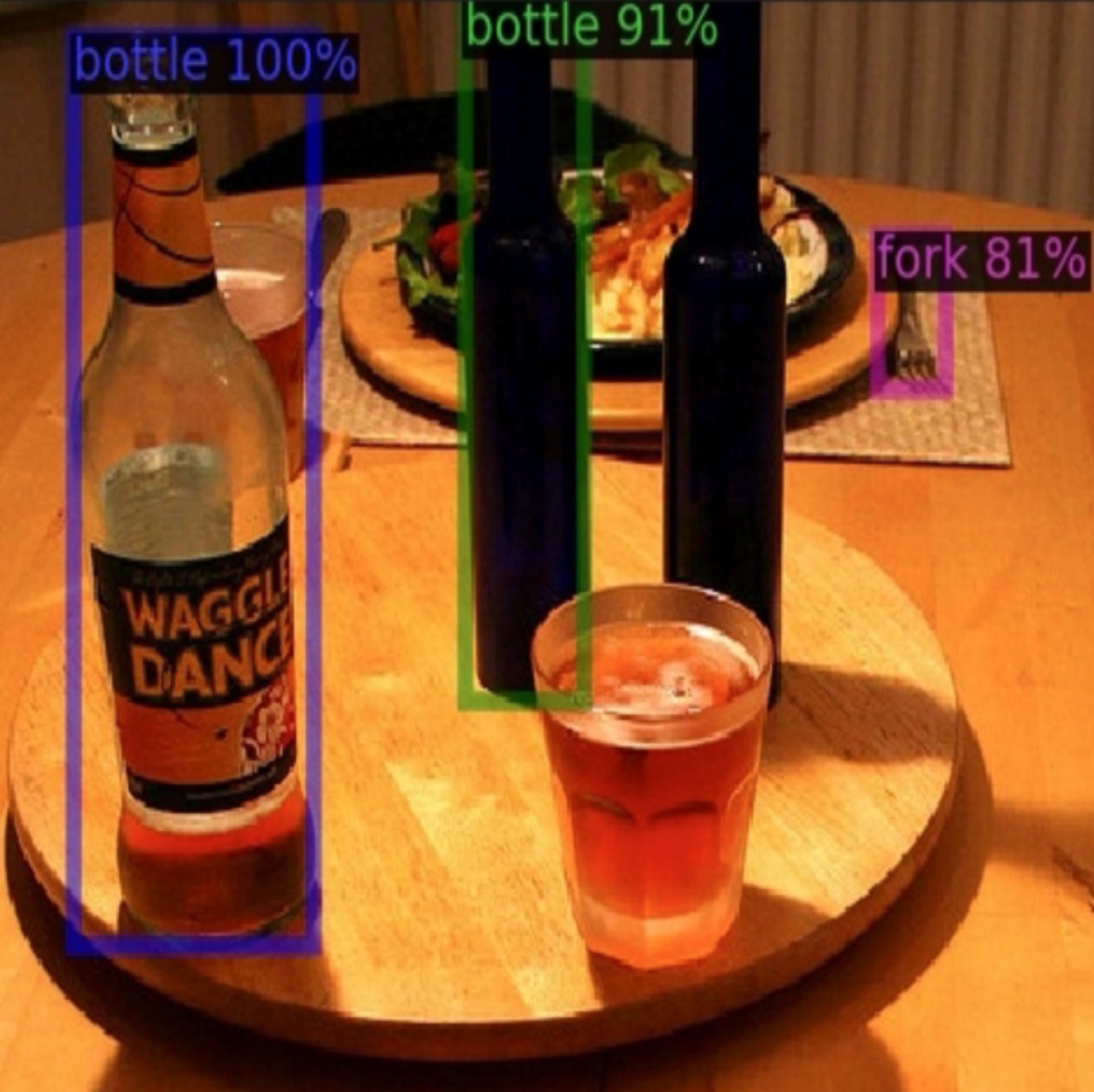} &
        \includegraphics[width=0.2\textwidth]{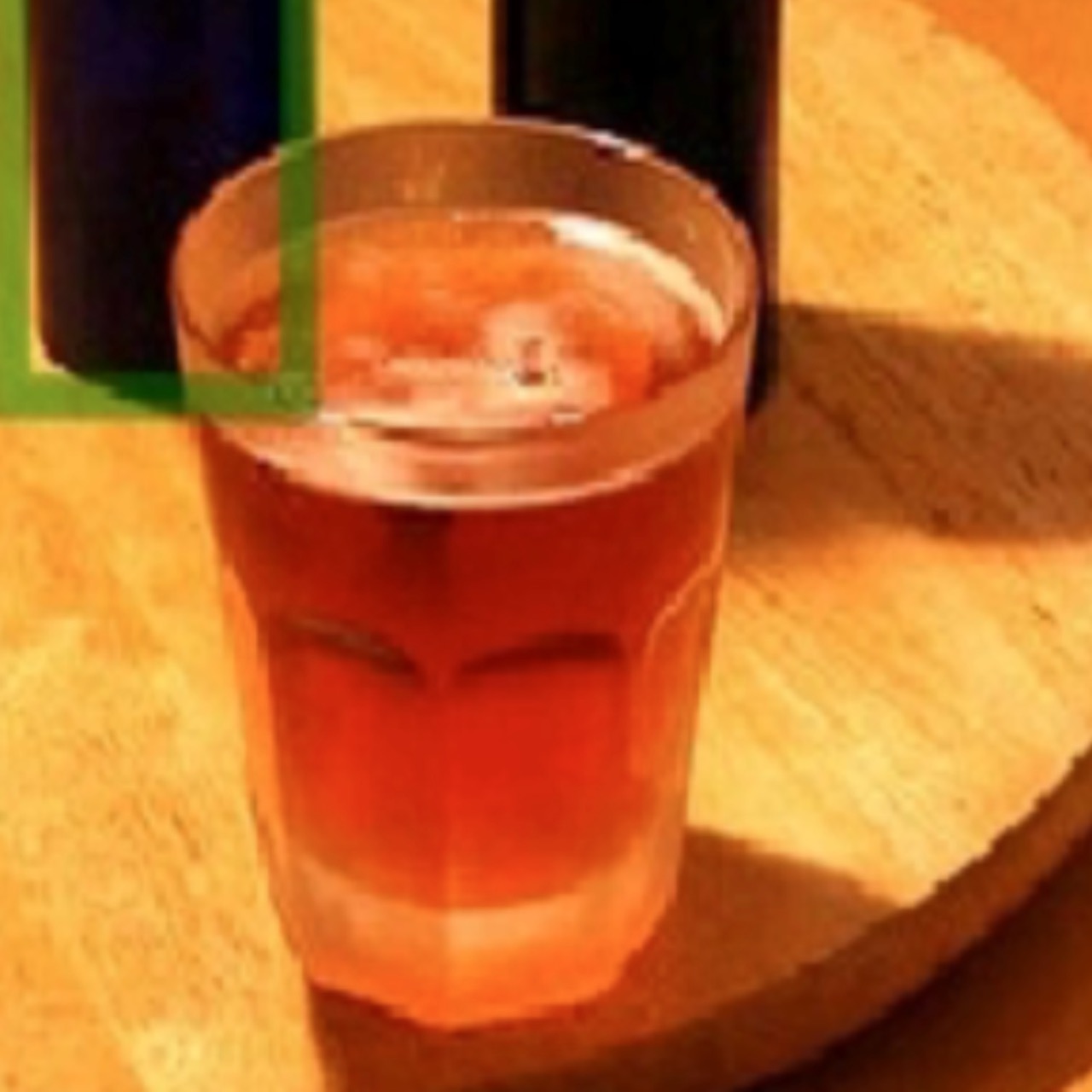} &
        \includegraphics[width=0.2\textwidth]{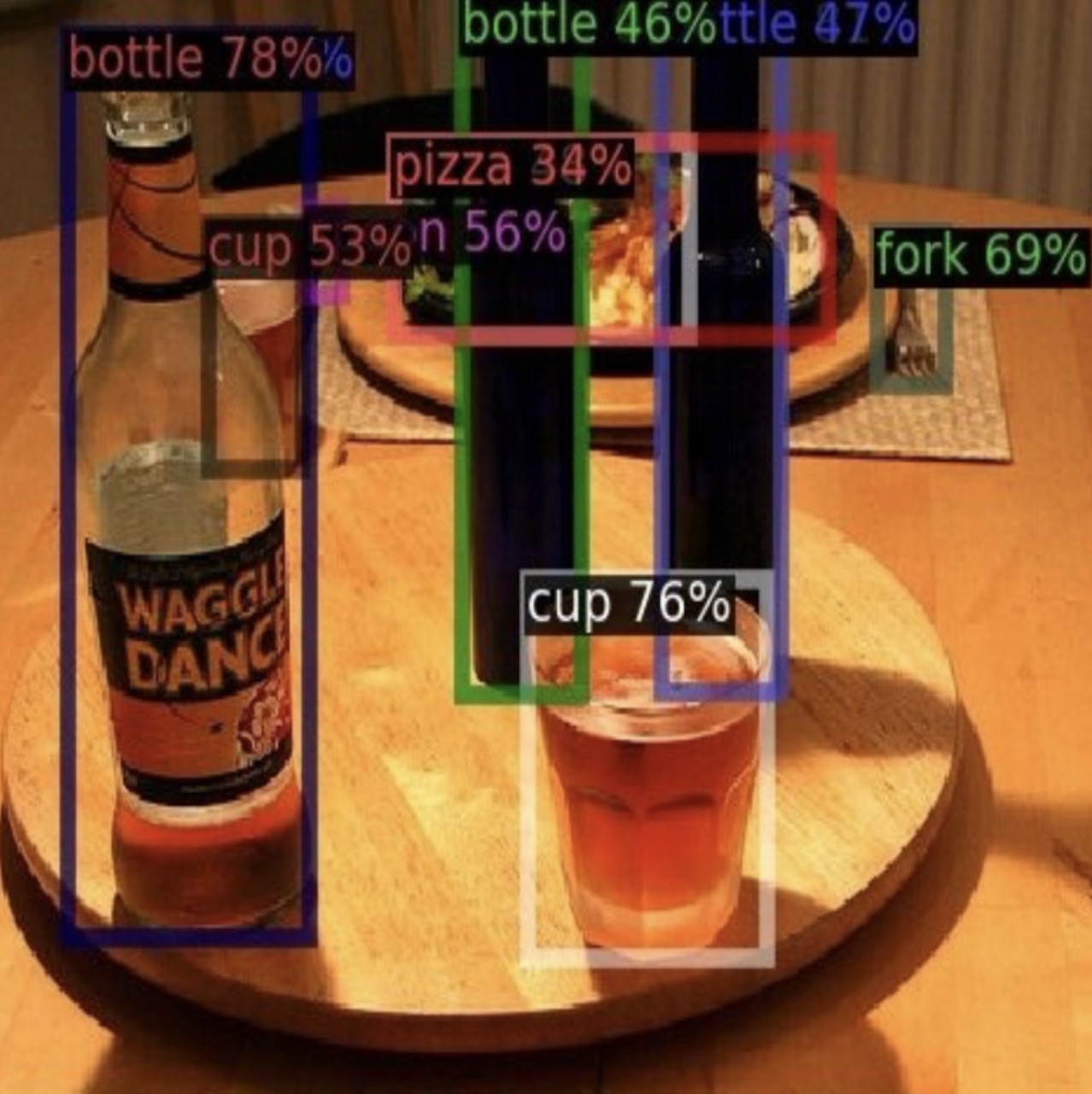} &
        \includegraphics[width=0.2\textwidth]{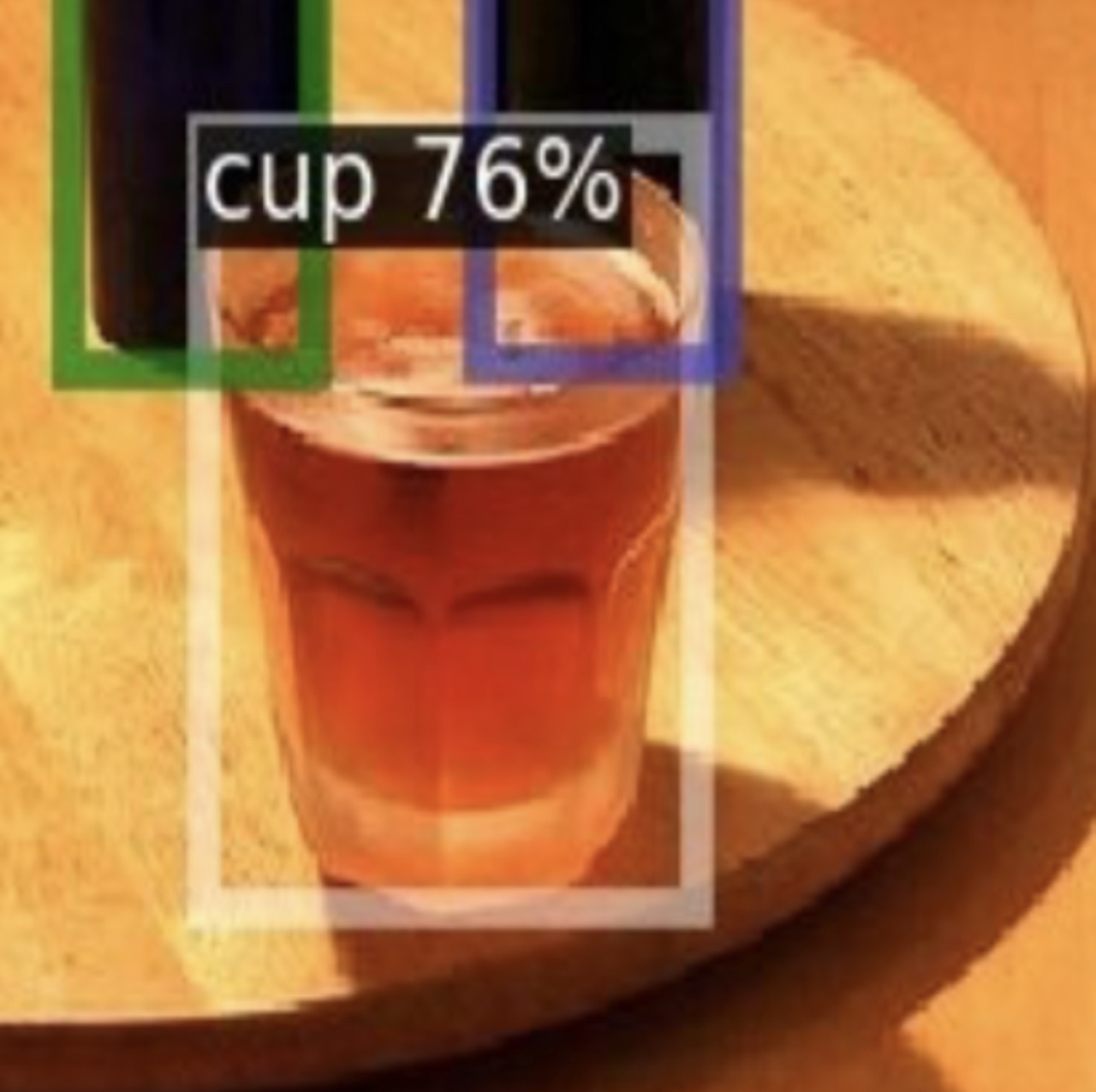}\\
        \includegraphics[width=0.2\textwidth]{ICPR_2026_self_supervised/figure/Raw/r3.jpeg} &
        \includegraphics[width=0.2\textwidth]{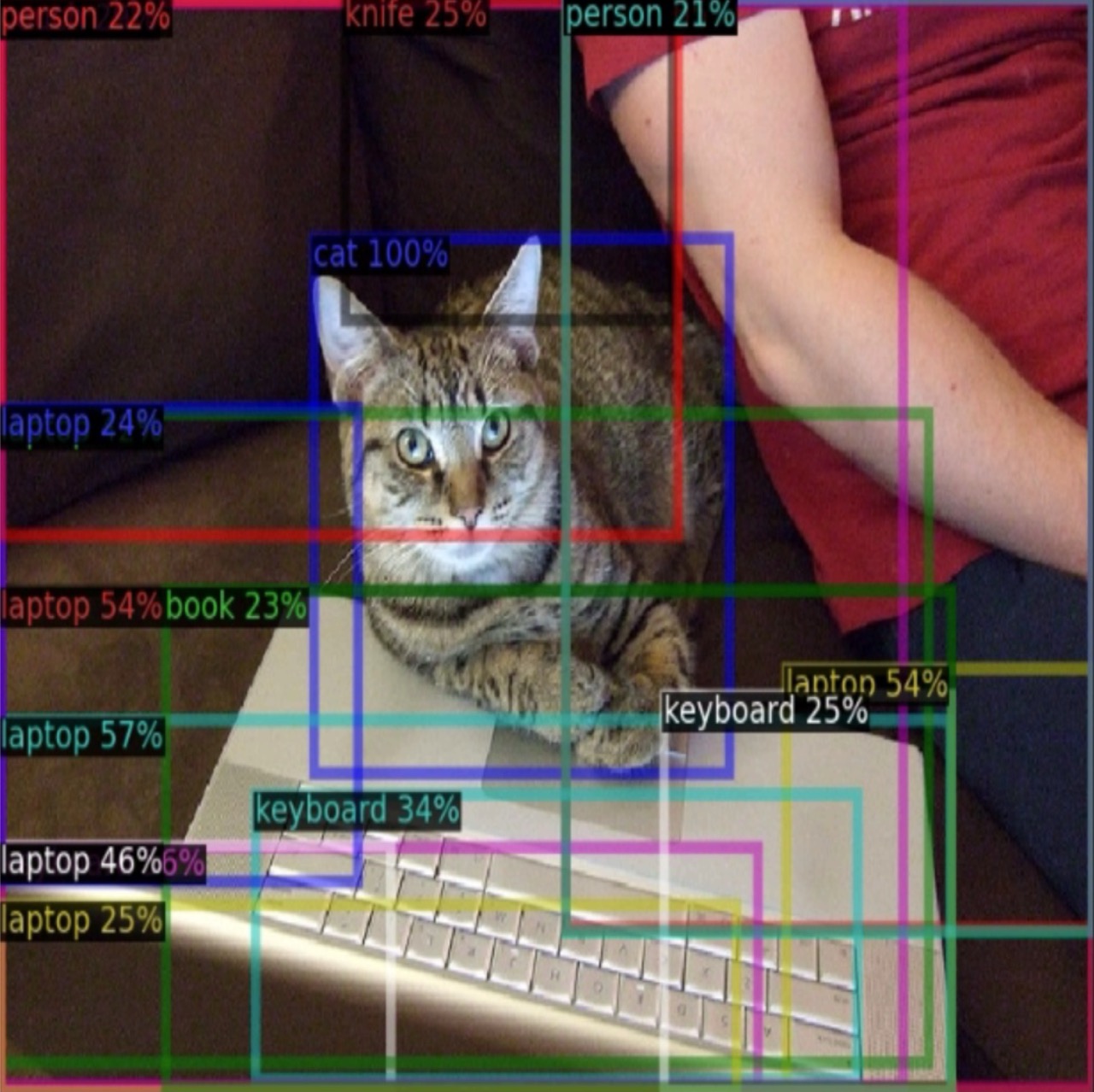} &
        \includegraphics[width=0.2\textwidth]{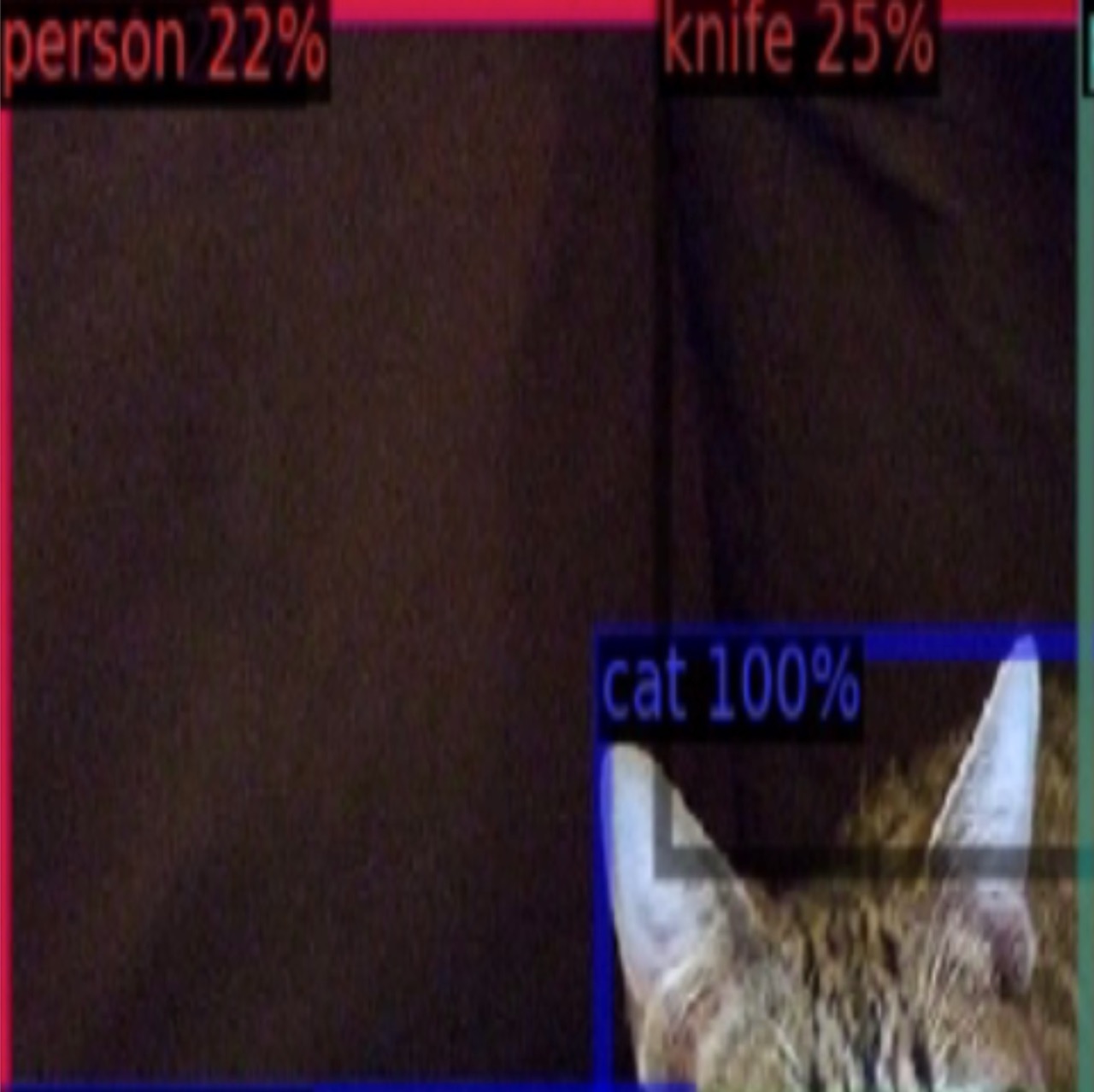} &
        \includegraphics[width=0.2\textwidth]{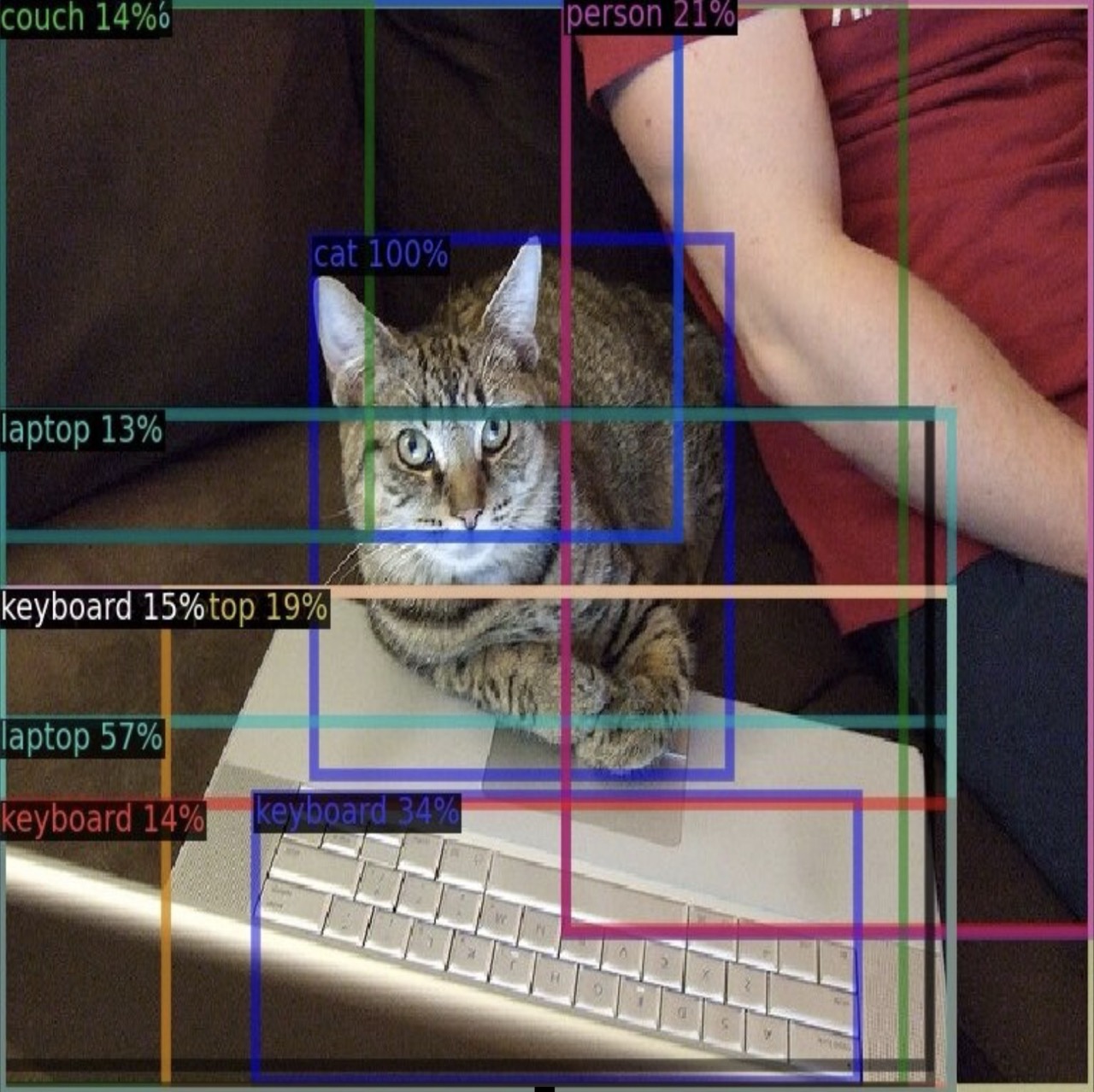} &
        \includegraphics[width=0.2\textwidth]{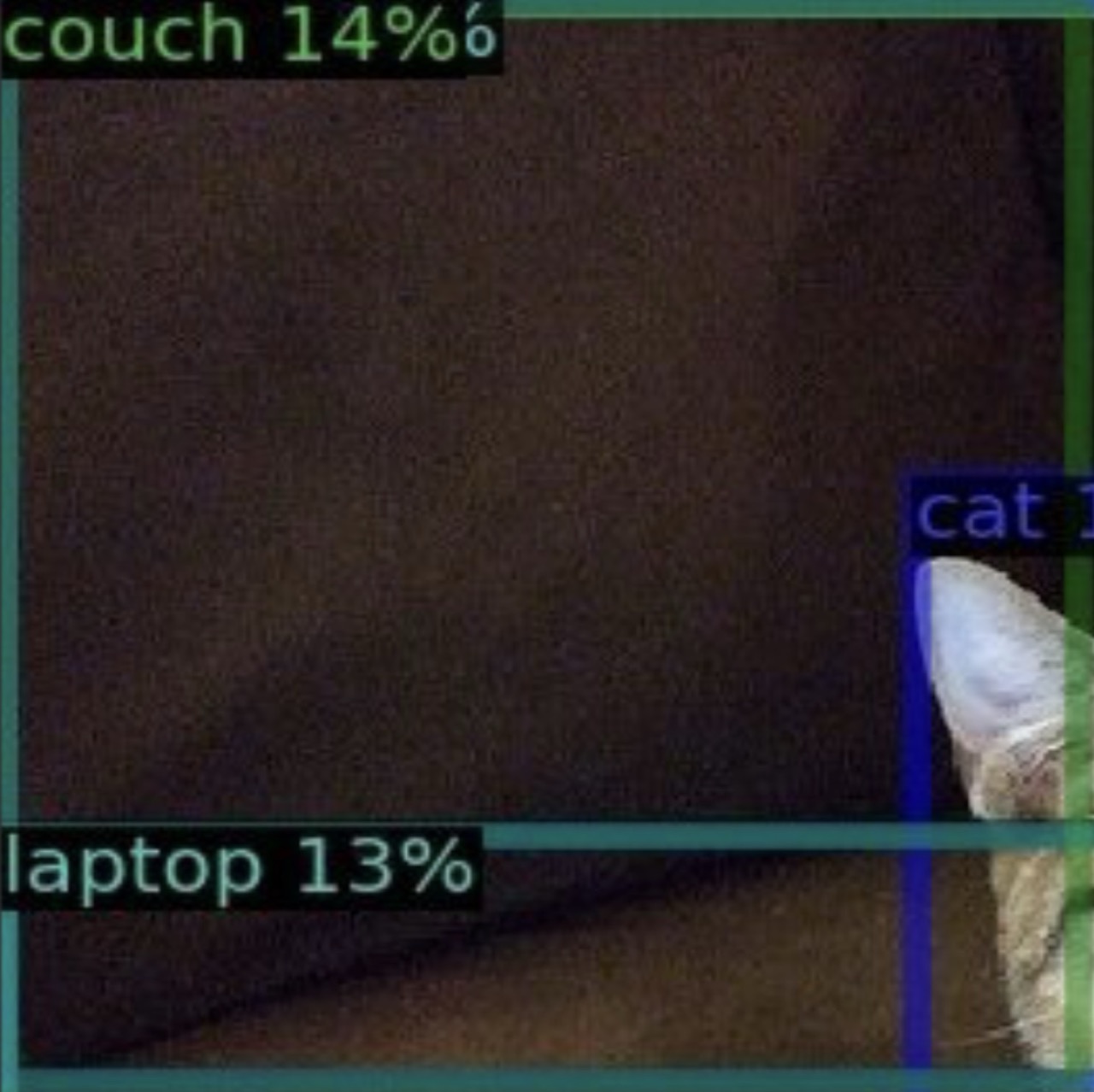}\\
                \includegraphics[width=0.2\textwidth]{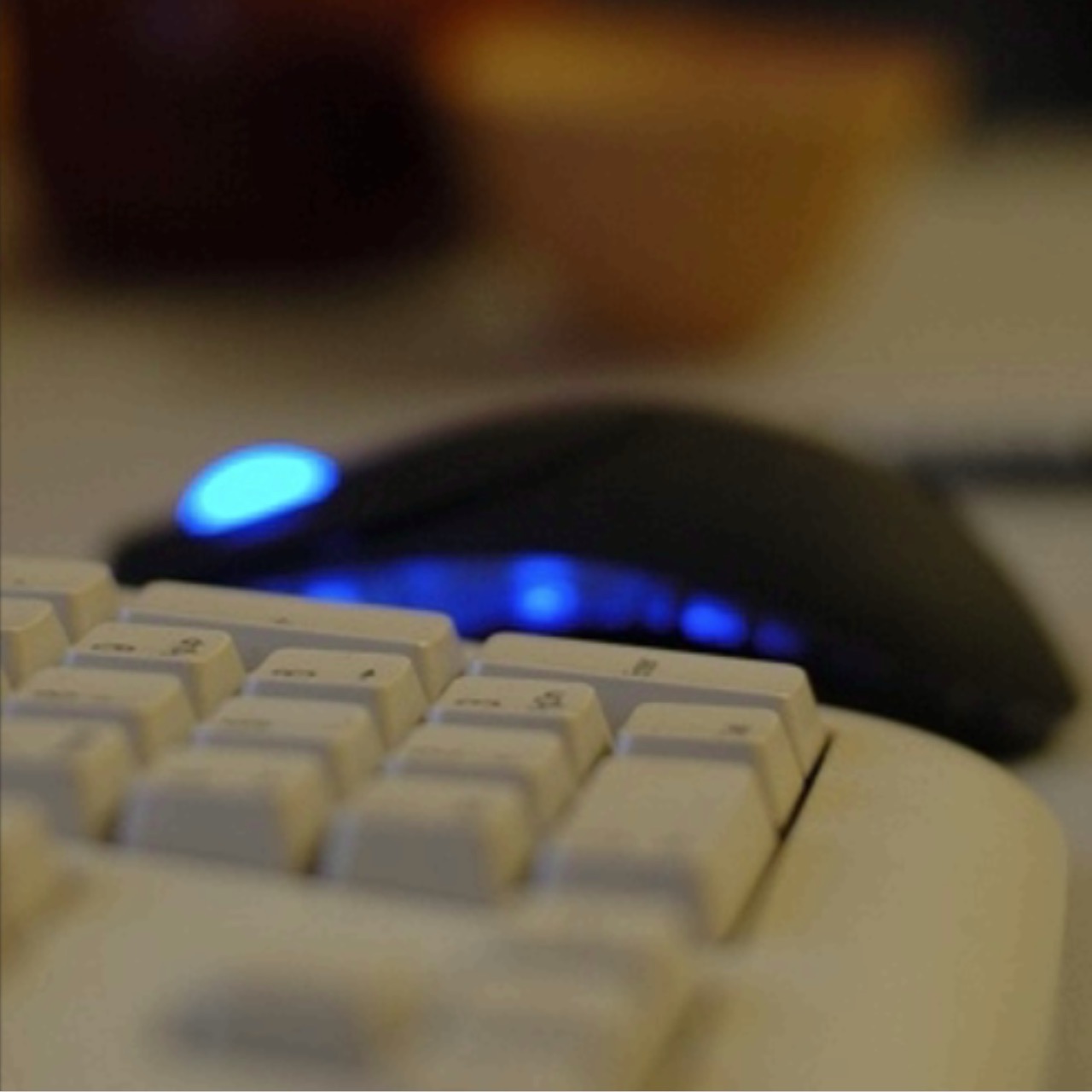} &
        \includegraphics[width=0.2\textwidth]{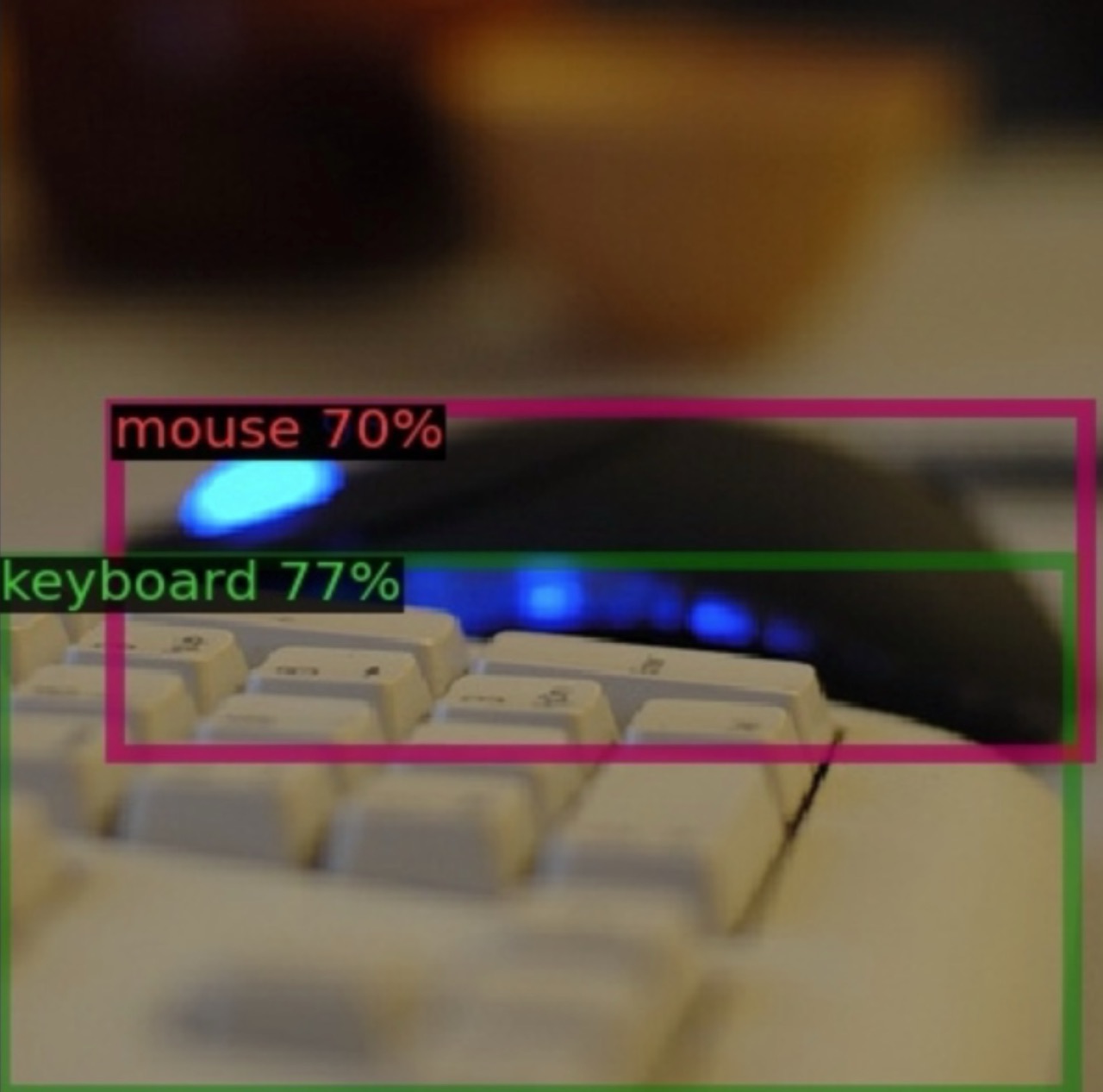} &
        \includegraphics[width=0.2\textwidth]{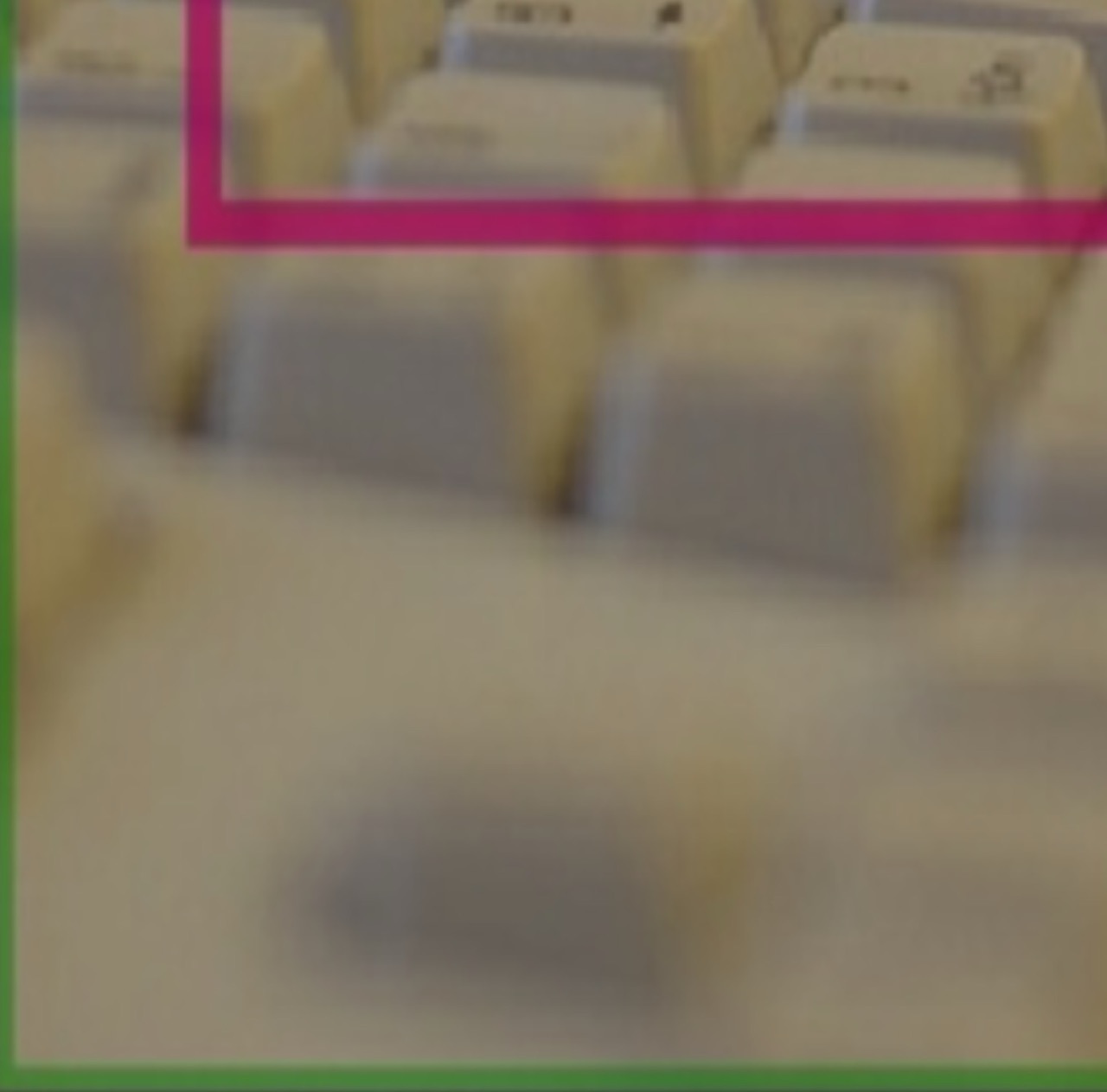} &
        \includegraphics[width=0.2\textwidth]{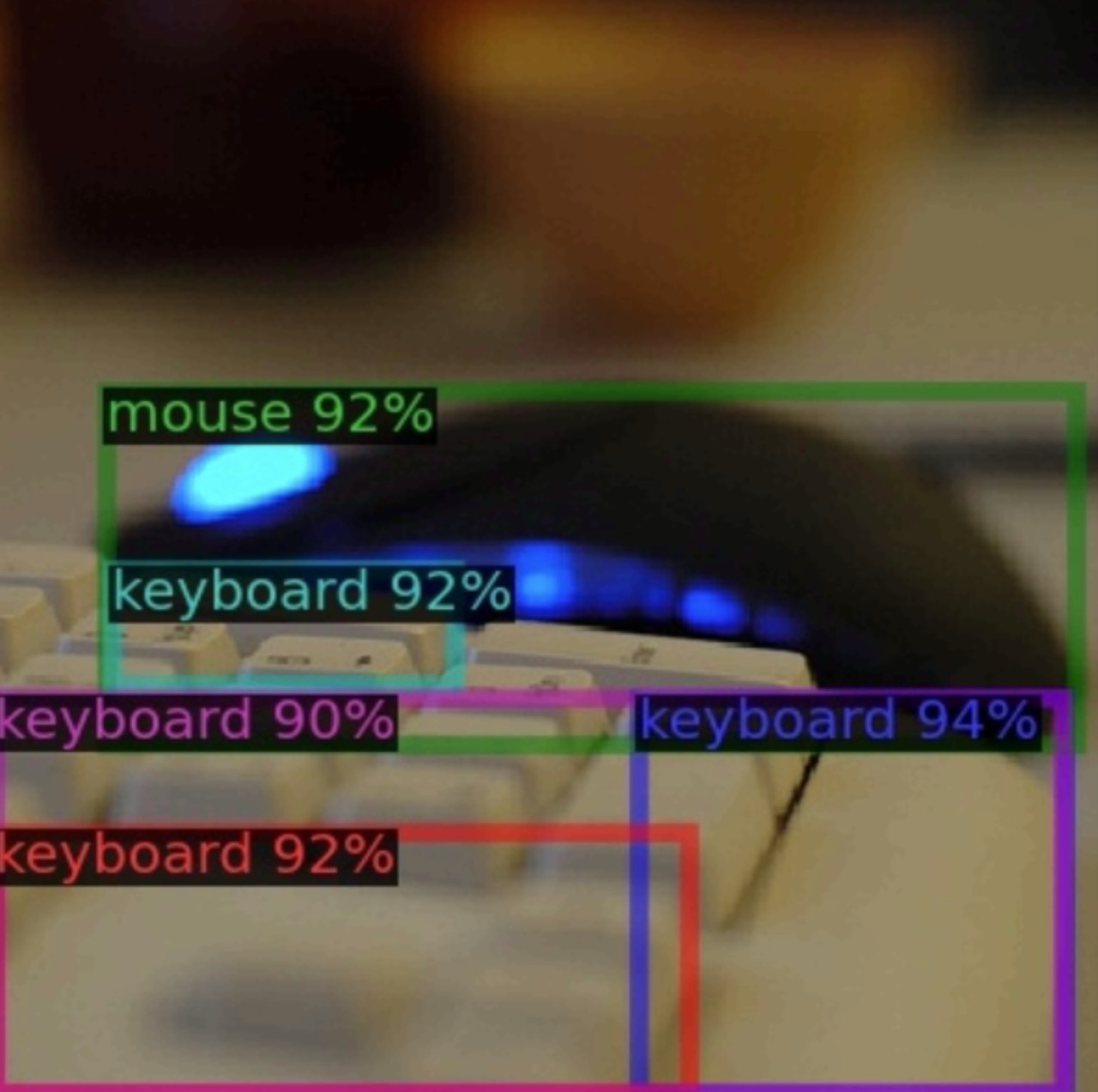} &
        \includegraphics[width=0.2\textwidth]{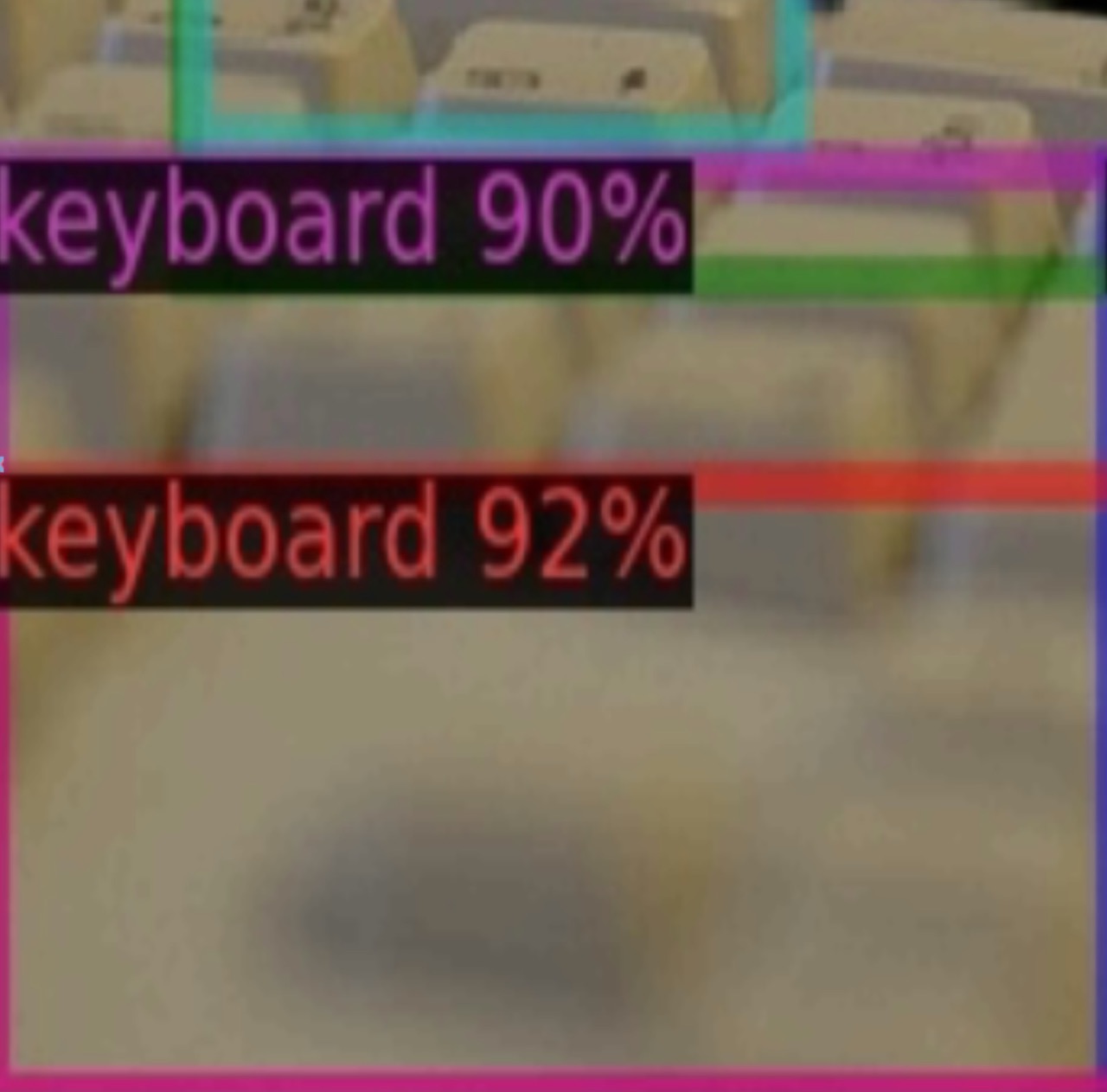}\\
        
        Input & CFM & CFM (crop) & Ours & Ours (crop) \\
    \end{tabular}}

    \caption{Sample results of Cooperative Foundational Models (CFM) and ours.}
    \label{fig:comp}
\end{figure}

Compared to conventional fine-tuning strategies that require multi-GPU training or full model updates, our approach achieves comparable or better open-vocabulary detection accuracy with less than 1M trainable parameters and a total training time under two hours.  
This shows that lightweight single-GPU fine-tuning can yield substantial benefits for cooperative large-model pipelines.

Figure~\ref{fig:comp} provides qualitative comparisons of detection results on COCO and LVIS validation images, which visually substantiate our method's contribution. For the sake of clarity, we only visualize the categories defined in the dataset and used for our evaluation. Compared to the baseline cooperative framework (CFM), our fine-tuned SigLIP backbone exhibits substantially improved region–text alignment, especially for challenging cases such as small, occluded, or semantically ambiguous objects. 

Specifically, in the first example, the baseline misclassifies a chair as a suitcase, likely because the cropped region containing a person's hand misleads the model into associating the scene with “holding an object.” In contrast, our adapted backbone correctly identifies the chair, demonstrating robust local feature understanding. In the second example, a cup on a table is missed entirely by the baseline but is accurately detected by our model, highlighting enhanced sensitivity to region-level details. The third row shows that the baseline incorrectly labels a couch as a person, a semantically inconsistent error that our method avoids, indicating better preservation of structural object semantics. Finally, in the fourth example, the baseline fails to produce high-confidence detections for a keyboard composed of multiple key regions, while our model successfully localizes all components, reflecting stronger fine-grained recognition ability.

These observations collectively confirm that our region-adapted backbone significantly reduces spurious detections and improves visual grounding in cluttered scenes, validating that targeted adaptation can enhance open-vocabulary reasoning without compromising the global semantic knowledge of the foundation model.

\subsection{Analysis}\label{subsec:ablation}
We conduct ablation experiments to examine the effects of three major design choices in DAT: dataset scale, bounding box generation strategy, and WiSE-FT integration. 

To analyze the effect of the dataset scale, we fine-tuned SigLIP using 100, 500, 1,000, and 5,000 source images, corresponding to approximately 8k, 40k, 80k, and 400k cropped regions, respectively.  
As shown in Table~\ref{tab:abl_dataset_scale}, performance peaks at 500–1,000 images and declines when scaled to 5,000 images. This reveals a critical \emph{saturation point} in region adaptation: VLMs benefit from sufficient but not excessive domain-specific data. Beyond this point, increased noise from pseudo-labels and gradient conflicts may outweigh the benefits of additional samples, highlighting the importance of data curation in self-supervised adaptation.

\input{ICPR_2026_self_supervised/tables/tab_scale}

We analyze the effect of region generation by comparing the crops generated by a self-supervised Mask R-CNN with those from ground-truth bounding boxes.  
Table~\ref{tab:abl-region} shows that self-supervised regions yield higher COCO performance (70.1 vs. 59.8 AP), while ground-truth boxes offer a slight edge on LVIS. This demonstrates the principle of \emph{distribution alignment}: training on proposals that match the inference-time distribution leads to stronger in-domain generalization, even if they are noisier than human annotations.

We further analyze the effect of WiSE-FT in balancing pre-trained knowledge and task-specific adaptation.  
Table~\ref{tab:abl-wiseft} shows that WiSE-FT ($\alpha = 0.5$) yields a substantial gain on LVIS (+1.7 AP) with negligible loss on COCO (–0.1 AP). This confirms that interpolating between original and adapted weights effectively preserves the VLM’s zero-shot capabilities while enabling robust region-level recognition—a key enabler for cross-dataset generalization.

\input{ICPR_2026_self_supervised/tables/tab_gen_wise}

In summary, these ablation studies validate DAT's core design: 1) moderate-scale self-supervised data avoids overfitting while enabling meaningful adaptation; 2) distribution-aligned region proposals can be more effective than ideal annotations for in-domain performance; and 3) weight-space interpolation maintains a practical balance between generalization and adaptation.

\textbf{Limitations.}
While DAT generally enhances the detection performance, its effectiveness depends on the quality of the base detector used for pseudo-label generation. This dependency explains the performance discrepancy between COCO and LVIS, i.e., pseudo-labels from a COCO-trained detector align well with the COCO validation distribution, enabling substantial gains, whereas the 1,123 long-tail categories in LVIS induce noisier pseudo-labels, limiting the capacity for improvement.
Consequently, DAT may require further fine-tuning or a better in-domain region proposer when applied to new domains.
Furthermore, as DAT operates within the current cooperative detection pipeline, the detection performance inevitably relies on the open-set detector.

%% file: ICPR_2026_self_supervised/tables/tab_coco.tex
\begin{table*}[t]
\centering
\caption{Results on COCO. Our method shows new state-of-the-art results, particularly on novel categories.}
\label{tab:main_results}
\setlength{\tabcolsep}{4pt} % 控制列间距
\renewcommand{\arraystretch}{1.0} % 控制行高
\fontsize{9pt}{12pt}\selectfont
\scalebox{0.865}{
\begin{tabular}{@{} >{\raggedright\arraybackslash}p{0.18\textwidth} 
                  >{\raggedright\arraybackslash}p{0.32\textwidth} 
                  >{\raggedright\arraybackslash}p{0.28\textwidth} 
                  c c c @{}}
\toprule[1.2pt]
\multicolumn{1}{l}{\textbf{Method}} & 
\multicolumn{1}{l}{\textbf{Pre-training}} & 
\multicolumn{1}{l}{\textbf{Detection Training Data}} & 
\multicolumn{3}{c}{\boldmath\textbf{AP\textsubscript{50}}} \\
\cmidrule(r){1-3} \cmidrule(l){4-6}
 & & & \textbf{Novel} & \textbf{Base} & \textbf{All} \\
\midrule
\rowcolor{gray!8} % 添加浅灰色背景区分范式
\multicolumn{6}{l}{\textbf{End-to-end models}} \\
\rowcolor{white}
OV-DETR~\cite{zang2022open} & -- & COCO, CLIP & 29.4 & 61.0 & 52.7 \\
DetCLIPv3~\cite{yao2024detclipv3} & FILIP, BLIP, GPT-4 & \mbox{O365, V3Det, GranuCap50M} & 54.7 & 42.8 & 46.9 \\
\addlinespace[0.3em] % 增加范式间间距
\rowcolor{gray!8}
\multicolumn{6}{l}{\textbf{Cooperative models}} \\
\rowcolor{white}
ViLD~\cite{gu2021open} & -- & COCO, CLIP & 27.6 & 59.5 & 51.3 \\
Detic~\cite{zhou2022detecting} & ImageNet-21K & COCO, IL, CC, CLIP & 27.8 & 47.1 & 45.0 \\
BARON~\cite{wu2023aligning} & SOCO dataset & COCO, CLIP & 34.0 & 60.4 & 53.5 \\
CORA~\cite{wu2023cora} & -- & COCO, CLIP & 35.1 & 35.5 & 35.4 \\
BARON~\cite{wu2023aligning} & SOCO, MAVL & COCO, CC, CLIP & 42.7 & 54.9 & 51.7 \\
CORA+~\cite{wu2023cora} & -- & COCO, CC, CLIP & 43.1 & 60.9 & 56.2 \\
CFM~\cite{bharadwaj2025enhancing} & GDINO, SAM, SigLIP & COCO & 50.3 & 49.8 & 49.9 \\
\midrule[1pt]
\rowcolor{blue!5} % 为我们的方法添加浅蓝色背景突出
\bfseries Ours & 
\bfseries \mbox{GDINO, SAM, SigLIP\textsubscript{ft}} & 
\bfseries COCO & 
\bfseries 70.1 & 
\bfseries 55.5 & 
\bfseries 59.3 \\
\bottomrule[1.2pt]
\end{tabular}
}
\vspace{0.5em}
\begin{minipage}{\textwidth}
\scriptsize
\raggedright
\textsuperscript{*} IL: ImageNet-Localization, CC: Conceptual Captions. \textbf{CLIP\textsubscript{ft}} denotes our proposed fine-tuning strategy. Bold numbers indicate best results within each column. Our method outperforms all prior cooperative and end-to-end approaches, achieving the highest novel-class AP\textsubscript{50} by a large margin.
\end{minipage}
\end{table*}

%% file: ICPR_2026_self_supervised/tables/tab_lvis.tex
\begin{table}[!t]
\centering
\caption{Open-vocabulary detection performance on LVIS v1.0 validation set. Our method outperforms existing approaches across all splits.}
\label{tab:lvis_results}
\setlength{\tabcolsep}{5pt}
\renewcommand{\arraystretch}{0.95}
\small
\begin{tabular}{@{} >{\raggedright\arraybackslash}p{0.25\textwidth} 
                  >{\centering\arraybackslash}p{0.18\textwidth} 
                  c c c @{}}
\toprule[1.2pt]
\multicolumn{1}{c}{\textbf{Method}} & 
\multicolumn{1}{c}{\textbf{VLM}} & 
\multicolumn{3}{c}{\boldmath\textbf{AP}} \\
\cmidrule(lr){1-2} \cmidrule(l){3-5}
 & & \textbf{Novel} & \textbf{Known} & \textbf{All} \\
\midrule
\rowcolor{gray!5}
\rowcolor{white}
K-Means~\cite{mcqueen1967some} & — & 0.20 & 17.77 & 1.55 \\
ORCA~\cite{cao2022open} & — & 0.49 & 20.57 & 2.03 \\
UNO~\cite{fini2021unified} & — & 0.61 & 21.09 & 2.18 \\
RNCDL~\cite{fomenko2022learning} & — & 5.42 & 25.00 & 6.92 \\
GDINO~\cite{liu2024grounding} & — & 13.47 & 37.13 & 15.30 \\

\addlinespace[0.3em]
\rowcolor{gray!5}

\rowcolor{white}
CFM (V2)~\cite{bharadwaj2025enhancing} & SigLIP & 17.42 & 42.08 & 19.33 \\

\midrule[1pt]
\rowcolor{blue!8}
\bfseries Ours & \bfseries SigLIP\textsubscript{ft} & 
\bfseries \cellcolor{blue!8}17.68 & 
\bfseries \cellcolor{blue!8}45.82 & 
\bfseries \cellcolor{blue!8}\boldmath 20.37 \\
\bottomrule[1.2pt]
\end{tabular}
\vspace{0.5em}
\begin{minipage}{\textwidth}
\scriptsize
\raggedright
\textsuperscript{*}\textbf{Novel}: 1123 rare categories; \textbf{Known}: 80 frequent/common categories. Our method establishes new state-of-the-art on LVIS open-vocabulary detection across all splits. Fine-tuning SigLIP through our DAT framework yields consistent improvements over the CFM baseline.
\end{minipage}
\end{table}

%% file: ICPR_2026_self_supervised/tables/tab_scale.tex
\begin{table}[t]
\centering
\caption{Ablation study on dataset scale for region adaptation. We show the correlation between the number of training region crops and open-vocabulary detection performance.}
\label{tab:abl_dataset_scale}
\setlength{\tabcolsep}{7pt}
\renewcommand{\arraystretch}{1.0}
\small
\begin{tabular}{@{} cc cc cc @{}}
\toprule[1.2pt]
\multicolumn{2}{c}{\textbf{Dataset Scale}} & \multicolumn{2}{c}{\textbf{COCO}} & \multicolumn{2}{c}{\textbf{LVIS}} \\
\cmidrule(lr){1-2} \cmidrule(lr){3-4} \cmidrule(l){5-6}
\# Images & \# Crops & \multicolumn{1}{c}{Novel AP} & $\Delta$ & \multicolumn{1}{c}{Novel AP} & $\Delta$ \\
\midrule
100       & $\sim$8k   & 53.87 & —      & 17.29 & —      \\
500       & $\sim$40k  & 70.03 & \textcolor{blue}{+16.16} & \textbf{17.68} & \textcolor{blue}{+0.39}  \\
\rowcolor{gray!5}
1,000     & $\sim$80k  & \textbf{70.11} & \textcolor{blue}{+16.24} & 17.56 & \textcolor{blue}{+0.27} \\
5,000     & $\sim$400k & 65.31 & \textcolor{blue}{+11.44} & 17.30 & \textcolor{blue}{+0.01}  \\
\bottomrule[1.2pt]
\end{tabular}
\vspace{0.5em}
\begin{minipage}{0.8\textwidth}
\scriptsize
\raggedright
\textsuperscript{*}\# Crops: the approximate number of detection crops. 
\end{minipage}
\end{table}

%% file: ICPR_2026_self_supervised/tables/tab_gen_wise.tex
\begin{table}[t]
\centering
\caption{Impact of region generation and WiSE-FT on fine-tuned results.}
\begin{subtable}{0.45\textwidth}
\centering
\caption{Ablation on region proposals.}
\label{tab:abl-region}
\adjustbox{max width=\linewidth}{
\begin{tabular}{@{}lcc@{}}
\toprule
Region Source & COCO AP & LVIS AP \\
\midrule
Mask R-CNN (self-supervised) & \textbf{70.1} & 17.7 \\
Ground-truth boxes & 59.8 & \textbf{18.1} \\
\bottomrule
\end{tabular}
}
\end{subtable}
\hfill
\begin{subtable}{0.48\textwidth}
\centering
\caption{Ablation on WiSE-FT .}
\label{tab:abl-wiseft}
\adjustbox{max width=\linewidth}{
\begin{tabular}{@{}lcc@{}}
\toprule
Setting & COCO Novel AP & LVIS Novel AP \\
\midrule
Without WiSE-FT & \textbf{70.2} & 16.0 \\
With WiSE-FT ($\alpha=0.5$) & 70.1 & \textbf{17.7} \\
\bottomrule
\end{tabular}
}
\end{subtable}
\label{tab:ablations}
\end{table}

%% file: ICPR_2026_self_supervised/content/conclusion.tex
\section{Conclusion and Future Work}
\label{sec:conclusion}

We presented DAT, which addresses the VLMs' limited ability in region-level
object recognition, enhancing cooperative model-based OVD. The key idea is to
perform PEFT of VLMs in a self-supervised manner using a pseudo-labeled
detection dataset obtained from the closed-set detector in the CFM pipeline.
Extensive experiments on COCO and LVIS demonstrate that DAT significantly
improves novel-class detection while maintaining strong base-class performance,
with minimal computational cost.
While DAT effectively bridges the global-local gap, its performance is
upper-bounded by the quality of pseudo-labels from the closed-set detector.
Future work includes online adaptation that enables
the model to continuously refine itself on high-confidence detections at
inference time, and validating our method on various VLM backbones to understand the generalization capability of DAT. 

\section{Acknowledgement}
\label{sec:ack}
This work was supported by the Korea Institute for Advancement of Technology (KIAT) grant (P0028485, GITCC).